\documentclass[11pt, a4paper, logo, acopyright]{googledeepmind}

\usepackage[utf8]{inputenc} % allow utf-8 input
\usepackage[T1]{fontenc}    % use 8-bit T1 fonts
\usepackage{hyperref}       % hyperlinks
\usepackage{cleveref}
\usepackage{cite}
\usepackage{url}            % simple URL typesetting
\usepackage{booktabs}       % professional-quality tables
\usepackage{amsfonts}       % blackboard math symbols
\usepackage{nicefrac}       % compact symbols for 1/2, etc.
\usepackage{microtype}      % microtypography
\usepackage{xcolor}         % colors

\usepackage{subcaption}
\usepackage{natbib}
\usepackage{tocloft}

\usepackage{amsmath,amsfonts,bm}

\def\eqref#1{equation~\ref{#1}}
\def\1{\bm{1}}

\DeclareMathAlphabet{\mathsfit}{\encodingdefault}{\sfdefault}{m}{sl}
\SetMathAlphabet{\mathsfit}{bold}{\encodingdefault}{\sfdefault}{bx}{n}

\usepackage{graphicx}
\usepackage{amsmath,bm}
\usepackage{amsthm}
\usepackage{wrapfig}
\usepackage{color, soul}
\usepackage{psfrag}
\usepackage{adjustbox}
\usepackage{xspace}
\usepackage{amssymb}
\usepackage{multirow}
\usepackage{afterpage}
\usepackage{colortbl}
\usepackage{float}
\usepackage{textcomp}
\usepackage{siunitx}
\usepackage{mdframed}
\usepackage[super]{nth}

\usepackage{enumitem}
\usepackage{bbding}
\usepackage{pifont}
\usepackage{bbm}
\usepackage{tcolorbox}
\usepackage{theoremref}
\colorlet{darkgreen}{green!65!black}
\colorlet{darkblue}{blue!75!black}
\colorlet{darkred}{red!80!black}
\definecolor{statistical}{HTML}{8c564b}
\definecolor{structural}{HTML}{0070C0}
\definecolor{semantic}{HTML}{008080}
\definecolor{yellow}{HTML}{f7c600}
\definecolor{lightblue}{HTML}{0071bc}
\definecolor{lightgreen}{HTML}{39b54a}
\definecolor{deemph}{gray}{0.55}

\definecolor{baselinecolor}{gray}{.95}
\definecolor{graycolor}{gray}{.95}

\newlength\savewidth
\newcolumntype{x}[1]{>{\centering\arraybackslash}p{#1pt}}
\newcolumntype{y}[1]{>{\raggedright\arraybackslash}p{#1pt}}
\newcolumntype{z}[1]{>{\raggedleft\arraybackslash}p{#1pt}}

\definecolor{textgreen}{RGB}{57, 172, 57}
\definecolor{textred}{RGB}{200, 10, 10}
\usepackage{pifont}

\title{CGM-JEPA: Learning Consistent Continuous Glucose Monitor Representations via Predictive Self-Supervised Pretraining}

\reportnumber{}

\author[1$*$]{Hada Melino Muhammad}
\author[1,2,$*$]{Zechen Li}
\author[1,$\dagger$]{Flora Salim}
\author[2,$\dagger$]{Ahmed A. Metwally}

\affil[1]{University of New South Wales}
\affil[2]{Google Research}
\affil[$*$]{Equal Contribution: hada\_melino.muhammad@unsw.edu.au, zechenl@google.com}
\affil[$\dagger$]{Correspondence: aametwally@google.com, flora.salim@unsw.edu.au}

\begin{abstract}
Continuous Glucose Monitoring (CGM) shows promise for detecting early metabolic subphenotypes such as insulin resistance (IR) and $\beta$-cell dysfunction, but its deployment for population-scale metabolic stratification faces two coupled problems. First, the same physiological state appears through multiple representational forms (raw CGM time series, sparse venous OGTT, distributional summaries such as Glucodensity), so a representation tied to a single view fails to transfer when deployment shifts the modality or setting. Second, baselines evaluated under such shifts perform inconsistently: each ranks well in some regimes and poorly in others. Both problems point to one remedy: representations that abstract away from any single view to capture higher-level temporal and distributional structure. We propose \textsc{CGM-JEPA}, a self-supervised predictive pretraining framework which predicts masked latent representations rather than reconstructing raw values, yielding abstraction that transfers across modalities. \textsc{X-CGM-JEPA} adds a masked Glucodensity cross-view objective that contributes complementary information from a distributional view. We pretrain on ${\sim}389$k unlabeled CGM readings from 228 subjects and evaluate on two clinical cohorts (Initial: $N\!=\!27$; Validation: $N\!=\!17$ in the public-release subset) across cohort generalization, venous-to-CGM transfer, and home CGM regimes, under a 20-iteration $\times$ 2-fold cross-validation protocol. \textsc{X-CGM-JEPA} ranks first or second on AUROC for both endpoints across all three evaluation regimes while no baseline stays in the top three, exceeding the strongest baseline by up to $+6.5$ AUROC points in cohort generalization and $+3.6$ points in venous-to-CGM transfer (paired Wilcoxon, $p < 0.001$). The cross-view design pays off where it should: in deployment settings under modality shift, \textsc{X-CGM-JEPA} matches mean AUROC while redistributing performance toward weaker subgroups (ethnicity AUROC gap shrinks $25$--$54\%$ under transfer); in the in-domain venous setting, where temporal context is sparse, the Glucodensity view lifts label-aware clustering (ARI $+39\%$, NMI $+40\%$ on the Initial cohort). Code, de-identified consented data, and pretrained weights are available at \url{https://github.com/cruiseresearchgroup/CGM-JEPA}.
\end{abstract}

\begin{document}

\maketitle

\newenvironment{Itemize}{
    \begin{itemize}[leftmargin=*]
    \setlength{\itemsep}{0pt}
    \setlength{\topsep}{0pt}
    \setlength{\partopsep}{0pt}
    \setlength{\parskip}{1pt}}
{\end{itemize}}
\setlength{\leftmargini}{9pt}

%%% Main Text: Introduction, Results, Discussion %%%
\section{Introduction}
\label{introduction}

\begin{figure*}[t]
\centering
\includegraphics[width=1\textwidth]{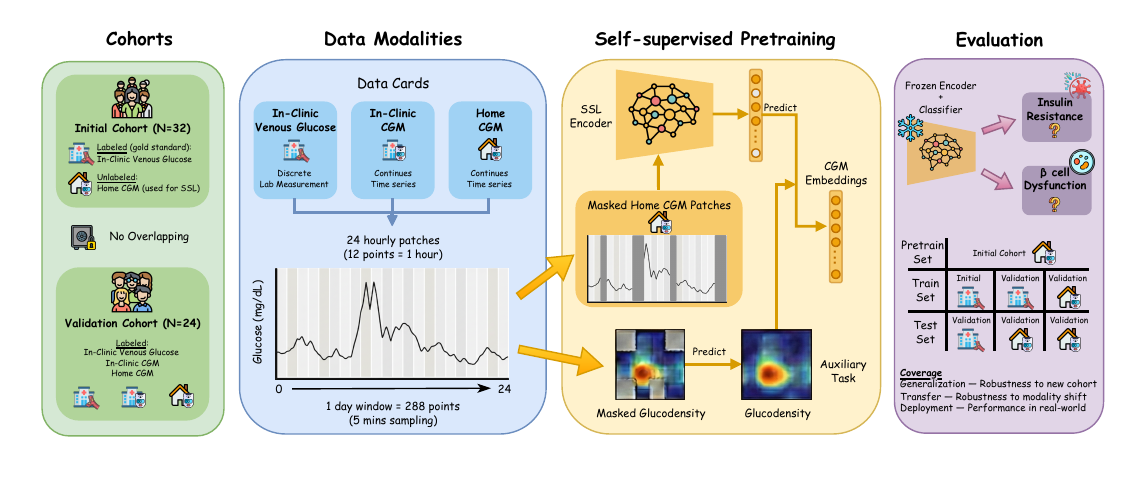} 
\caption{Study overview. Two cohorts with venous (gold-standard) and CGM measurements from controlled (In-Clinic) and free-living (Home) settings. We pretrain on unlabeled Home CGM (Initial cohort plus the public Col\'as cohort, with validation subjects excluded), then evaluate metabolic dysfunction prediction under cohort generalization, venous-to-CGM transfer, and home CGM settings.}
\label{fig:overview}
\vspace{-0.5em}
\end{figure*}

Continuous glucose monitoring (CGM) enables dense, continuous measurements of glucose dynamics and is increasingly adopted in normoglycemic and prediabetic populations~\cite{10.1145/3097983.3098068, sergazinov2024glucobenchcuratedlistcontinuous, Park2025LifestyleT2DSubphenotypes}. Beyond tracking average glucose levels, a central clinical goal is to uncover \emph{latent metabolic dysfunctions} that may underlie superficially similar glucose trajectories. In particular, insulin resistance and $\beta$-cell dysfunction represent two distinct physiological mechanisms on the path toward Type~2 Diabetes, yet they can produce overlapping CGM patterns depending on diet, activity, and daily routines. Accurately distinguishing these dysfunctions from CGM would enable earlier risk stratification and personalized intervention.

Despite this promise, deploying CGM-based subphenotype prediction at population scale faces two coupled problems. The first is a \emph{multi-view representation problem}: the same physiological state appears through multiple representational forms, including raw CGM time series, sparse venous OGTT measurements, and distributional summaries such as Glucodensity. Each view captures different aspects of glucose physiology, and a representation tied to any single view tends to fail when deployment shifts the modality (venous to CGM), the setting (controlled to free-living), or the cohort. The second is a \emph{consistency problem}: methods evaluated under such shifts perform inconsistently, with each baseline ranking well in some regimes and poorly in others, leaving no reliable choice for end-to-end CGM deployment. Together, these problems mean that label scarcity (gold-standard venous OGTT labels are costly and invasive~\cite{metwally2025prediction}) compounds with view fragility, and a method that performs well in one regime offers no guarantee in another.

Both problems point to a single underlying remedy: representations that abstract away from any specific view to capture higher-level temporal and distributional structure that is invariant across modalities and settings. Prior work on CGM modeling~\cite{metwally2025prediction, metwally2025usecontinuousglucosemonitoring, Wu2025GlycemicCarbsPhysiology} often relies on handcrafted feature pipelines (e.g., summary statistics or engineered glycemic indices) that operate on a single view and may not be stable under cohort and setting shifts. Recent time-series foundation models~\cite{goswami2024moment, ansari2024chronos, feofanov2025mantislightweightcalibratedfoundation, li-etal-2025-sensorllm, zhang2025sensorlmlearninglanguagewearable, 10.5555/3692070.3692474, Luo2024ALS} and self-supervised learning approaches~\cite{9157636, 10.5555/3524938.3525087, jaiswal2021surveycontrastiveselfsupervisedlearning, zhou-etal-2021-self-supervised, chen2025comodocrossmodalvideotoimudistillation} reduce reliance on labels, but most are evaluated within a single domain or modality and do not target the clinically realistic regime where supervision and deployment differ in both modality and setting. Many also rely on raw-signal reconstruction or contrastive augmentation objectives, which tie representations to surface signal properties rather than higher-level abstraction. This motivates the question we study: \emph{can we learn CGM representations that abstract beyond any single view and deliver consistent performance across the deployment regimes that matter for population-scale metabolic stratification?}

We address this question using a two-cohort design with complementary modality availability, enabling systematic evaluation across realistic deployment conditions. The \emph{Initial cohort} provides paired venous and home CGM measurements for one set of subjects, while the \emph{Validation cohort} provides labeled venous measurements alongside CGM collected in both controlled and home settings, supporting cross-cohort and cross-modality transfer evaluation. We pretrain representations using unlabeled home CGM from the Initial cohort and the publicly available Col\'as cohort~\cite{colas2019detrended}, with all validation-cohort subjects excluded from pretraining to prevent leakage. Figure~\ref{fig:overview} summarizes the cohorts, modalities, SSL pretraining pipeline, and downstream tasks. Our evaluation focuses on two binary outcomes (insulin resistance and $\beta$-cell dysfunction) under three clinically motivated regimes that span both deployment paths: cohort generalization, venous-to-CGM transfer, and real-world home CGM.

To deliver abstraction-first representations, we adopt a Joint Embedding Predictive Architecture~\cite{assran2023self, assran2023selfsupervisedlearningimagesjointembedding, weimann2025self, chen2025vl, dong2024brain}. JEPA's defining choice is to predict in latent space rather than reconstruct raw values, which encourages the encoder to capture higher-level structure that survives view changes rather than memorizing surface signal properties. We instantiate this as \textsc{CGM-JEPA}, a masked representation prediction objective for 1-day CGM windows that predicts latent representations of masked temporal patches~\cite{Yuqietal-2023-PatchTST} from the visible context. Building on this, \textsc{X-CGM-JEPA} extends the abstraction principle from a single view to multiple views: it adds an auxiliary predictive objective that predicts masked Glucodensity representations from the CGM context embedding, deliberately injecting complementary high-level information from a distributional view of the same window. Conceptually, \textsc{X-CGM-JEPA} treats abstraction as \emph{additive}: when one view leaves gaps, the complementary view fills them; when both views agree, they reinforce each other.

Across a broad set of baselines, including classical unsupervised projection~\cite{shlens2014tutorialprincipalcomponentanalysis, yue2022ts2vecuniversalrepresentationtime}, CGM-specific foundation models~\cite{lutsker2025gluformer}, and modern time-series foundation models~\cite{goswami2024moment, feofanov2025mantislightweightcalibratedfoundation}, the \textsc{CGM-JEPA} family delivers the consistency baselines lack. \textsc{X-CGM-JEPA} ranks first or second on AUROC for both endpoints (IR and $\beta$-cell dysfunction) across all three evaluation regimes (in-domain venous, in-domain home CGM, and venous-to-CGM transfer), while no baseline stays in the top three throughout, with each strong baseline winning some regimes and losing others. The cross-view design then pays off in two distinct regimes that map directly to the additive-abstraction principle. In deployment settings under modality shift, \textsc{X-CGM-JEPA} matches mean AUROC against \textsc{CGM-JEPA} but redistributes performance toward weaker demographic subgroups, indicating that the additional distributional view stabilizes representations under shift. In the in-domain venous setting, where the temporal context is sparse compared to continuous CGM, the Glucodensity view contributes additive structure that lifts label-aware clustering agreement, supporting the broader claim that complementary high-level information from a different view genuinely augments a temporal-only representation. To our knowledge, this is the first JEPA-style masked representation prediction framework instantiated for CGM time-series. Our contributions are threefold:

\begin{itemize}
    \item \textbf{Deployment-oriented problem formulation and protocol.} We formalize CGM subphenotype prediction as a two-path deployment problem (in-domain home CGM and venous-to-CGM transfer), under multi-view representation pressure and across three clinically motivated regimes evaluated within a unified, variance-controlled protocol (20-iteration $\times$ 2-fold subject-level cross-validation).
    \item \textbf{Abstraction-first CGM self-supervision.} We introduce \textsc{CGM-JEPA}, a JEPA-style masked latent prediction framework that operationalizes representation abstraction for CGM, yielding embeddings that consistently rank in the top two across all evaluation regimes while no baseline does.
    \item \textbf{Additive cross-view abstraction.} We propose \textsc{X-CGM-JEPA}, which extends the abstraction principle by predicting masked Glucodensity latents alongside CGM, contributing complementary high-level information from a distributional view. The cross-view design yields regime-specific value: subgroup-robust performance under deployment shift, and improved label-aware structure when the temporal view is data-thin.
\end{itemize}
\section{Results}
\label{results}

We evaluate two binary outcomes (insulin resistance, IR; $\beta$-cell dysfunction) under three clinically motivated regimes that span both deployment paths introduced in Section~\ref{introduction}: (i) in-domain home CGM, (ii) venous-to-CGM transfer, and (iii) in-domain venous (cohort generalization). All methods follow an identical evaluation protocol: subject-level stratified 2-fold cross-validation repeated over 20 random iterations (40 evaluations per cell), with frozen embeddings probed by Logistic Regression. We report mean$\pm$std AUROC, F1-score, and PRAUC averaged across runs; best and second-best are highlighted in bold and underlined respectively. All \textsc{X-CGM-JEPA} results in this section use the fixed auxiliary weight $\lambda{=}1$.

Our headline findings are twofold. First, the \textsc{CGM-JEPA} family delivers \emph{consistency} that no baseline matches: across every (endpoint $\times$ regime) cell, \textsc{X-CGM-JEPA} ranks first or second on AUROC, while every baseline drops to rank three or worse in at least one cell. Pooled across 108 paired comparisons (3 metrics $\times$ 6 endpoint-regime cells $\times$ 6 baselines), \textsc{CGM-JEPA} wins $101/108$ and \textsc{X-CGM-JEPA} wins $103/108$ (paired Wilcoxon, $p<0.001$ for both). Second, \textsc{X-CGM-JEPA}'s additive cross-view design yields its clearest distinct contribution in two specific regimes that map directly to the abstraction-as-additive principle introduced in Section~\ref{introduction}: in deployment under modality shift, where Glucodensity stabilizes performance across demographic subgroups (Section~\ref{subsec:subgroup-redistribution}); and in the in-domain venous setting, where the temporal context is sparse and the distributional view contributes complementary structure that lifts label-aware clustering (Section~\ref{subsec:representation-analysis}).

\subsection{In-Domain Home CGM}
\label{subsec:home-cgm}

We first evaluate the deployment-relevant in-domain home CGM regime: training and evaluation both use free-living home CGM within the validation cohort. This regime most closely matches population-scale deployment, where wearable CGM is collected under unconstrained daily-life conditions and exhibits behavioral variability, sensor noise, and missingness that the in-clinic measurements do not.

\begin{table}[t]
\centering
\caption{In-Domain Home CGM: $\beta$-cell Dysfunction.}
\label{tab:home_cgm_in_domain_beta}
\footnotesize
\begin{tabular}{lccc}
\hline
\textbf{Method} & \textbf{AUROC} & \textbf{F1-score} & \textbf{PRAUC} \\
\hline
PCA & $0.925 \pm 0.069$ & $0.760 \pm 0.122$ & $0.909 \pm 0.081$ \\
\textsc{GluFormer} & $0.857 \pm 0.121$ & $0.649 \pm 0.111$ & $0.809 \pm 0.160$ \\
TS2Vec & $0.854 \pm 0.140$ & $0.738 \pm 0.118$ & $0.880 \pm 0.107$ \\
$\text{MOMENT}_{\text{Small}}$ & $0.849 \pm 0.094$ & $0.591 \pm 0.164$ & $0.834 \pm 0.096$ \\
$\text{MOMENT}_{\text{Large}}$ & $0.652 \pm 0.147$ & $0.511 \pm 0.172$ & $0.649 \pm 0.141$ \\
Mantis & $0.758 \pm 0.098$ & $0.573 \pm 0.175$ & $0.777 \pm 0.089$ \\
\hline
CGM-JEPA & \underline{$0.932 \pm 0.077$} & \underline{$0.806 \pm 0.166$} & \underline{$0.931 \pm 0.076$} \\
X-CGM-JEPA & $\mathbf{0.946 \pm 0.063}$ & $\mathbf{0.811 \pm 0.169}$ & $\mathbf{0.941 \pm 0.066}$ \\
\hline
\end{tabular}
\vspace{-0.5em}
\end{table}

\begin{table}[t]
\centering
\caption{In-Domain Home CGM: Insulin Resistance.}
\label{tab:home_cgm_in_domain_ir}
\footnotesize
\begin{tabular}{lccc}
\hline
\textbf{Method} & \textbf{AUROC} & \textbf{F1-score} & \textbf{PRAUC} \\
\hline
PCA & $0.821 \pm 0.133$ & $0.694 \pm 0.151$ & $0.842 \pm 0.114$ \\
\textsc{GluFormer} & $\mathbf{0.889 \pm 0.103}$ & $0.701 \pm 0.105$ & \underline{$0.880 \pm 0.114$} \\
TS2Vec & $0.848 \pm 0.130$ & $0.676 \pm 0.148$ & $0.868 \pm 0.101$ \\
$\text{MOMENT}_{\text{Small}}$ & $0.819 \pm 0.101$ & $0.640 \pm 0.148$ & $0.829 \pm 0.089$ \\
$\text{MOMENT}_{\text{Large}}$ & $0.605 \pm 0.127$ & $0.537 \pm 0.143$ & $0.651 \pm 0.124$ \\
Mantis & $0.690 \pm 0.127$ & $0.570 \pm 0.128$ & $0.759 \pm 0.093$ \\
\hline
CGM-JEPA & $0.842 \pm 0.118$ & \underline{$0.746 \pm 0.168$} & $0.872 \pm 0.091$ \\
X-CGM-JEPA & \underline{$0.857 \pm 0.112$} & $\mathbf{0.754 \pm 0.161}$ & $\mathbf{0.883 \pm 0.086}$ \\
\hline
\end{tabular}
\vspace{-0.5em}
\end{table}

\paragraph{$\beta$-cell Dysfunction.}
Table~\ref{tab:home_cgm_in_domain_beta} shows that \textsc{X-CGM-JEPA} achieves the best AUROC, F1, and PRAUC on $\beta$-cell prediction, with \textsc{CGM-JEPA} second across all three metrics. Compared to the strongest baseline (PCA), \textsc{X-CGM-JEPA} improves AUROC by \textbf{+2.1}~pp, F1 by \textbf{+5.1}~pp, and PRAUC by \textbf{+3.2}~pp. The two JEPA variants form a tight pair, with \textsc{X-CGM-JEPA} marginally ahead on AUROC ($+0.014$) and PRAUC ($+0.010$). Two further patterns are notable. First, only the JEPA variants exceed F1~$=0.80$ ($0.811$ and $0.806$), while the next best baseline (PCA) reaches $0.760$, a $5$-point gap that translates directly to operating-point performance for screening deployment. Second, \textsc{X-CGM-JEPA} attains the lowest fold-to-fold AUROC standard deviation among all methods ($0.063$ vs.\ PCA $0.069$), indicating that the JEPA family's consistency extends from cross-regime stability to within-regime robustness.

\paragraph{Insulin Resistance.}
For insulin resistance (Table~\ref{tab:home_cgm_in_domain_ir}), \textsc{GluFormer} achieves the best AUROC ($0.889$), edging \textsc{X-CGM-JEPA} by $3.2$~pp. This is the single endpoint--regime cell where a baseline outranks the JEPA family on AUROC. The advantage is regime-specific: \textsc{GluFormer} trails the JEPA family by $7$ to $34$ AUROC points across the other five endpoint--regime cells, with near-random performance ($0.530$) on Venous-to-CGM IR transfer (Section~\ref{subsec:transfer}). \textsc{X-CGM-JEPA} stays competitive ($0.857$, second) and achieves the best F1 ($0.754$) and PRAUC ($0.883$), indicating a more favorable threshold-dependent profile despite the AUROC gap. Across both endpoints, the JEPA family stays in the top two; no baseline does.

\subsection{Cross-Modality Transfer}
\label{subsec:transfer}

We next evaluate the cross-modality deployment path: classifiers are trained on venous-supervised embeddings and tested on home-CGM embeddings within the validation cohort. This regime mirrors a practical screening scenario in which gold-standard labels come from clinical venous assays but inference at scale is performed on consumer-grade wearable CGM, exposing methods to a simultaneous modality shift (venous to CGM) and setting shift (controlled to free-living).

\begin{table}[t]
\centering
\caption{Venous $\to$ Home CGM: $\beta$-cell Dysfunction.}
\label{tab:venous_to_home_cgm_beta}
\footnotesize
\begin{tabular}{lccc}
\hline
\textbf{Method} & \textbf{AUROC} & \textbf{F1-score} & \textbf{PRAUC} \\
\hline
PCA & $0.927 \pm 0.074$ & $0.673 \pm 0.103$ & $0.911 \pm 0.088$ \\
\textsc{GluFormer} & $0.801 \pm 0.129$ & $0.634 \pm 0.104$ & $0.756 \pm 0.160$ \\
TS2Vec & $0.805 \pm 0.131$ & $0.691 \pm 0.144$ & $0.754 \pm 0.167$ \\
$\text{MOMENT}_{\text{Small}}$ & $0.736 \pm 0.192$ & $0.609 \pm 0.157$ & $0.751 \pm 0.173$ \\
$\text{MOMENT}_{\text{Large}}$ & $0.606 \pm 0.229$ & $0.431 \pm 0.212$ & $0.615 \pm 0.228$ \\
Mantis & $0.821 \pm 0.102$ & $0.661 \pm 0.154$ & $0.826 \pm 0.105$ \\
\hline
CGM-JEPA & \underline{$0.946 \pm 0.063$} & \underline{$0.697 \pm 0.094$} & \underline{$0.941 \pm 0.066$} \\
X-CGM-JEPA & $\mathbf{0.949 \pm 0.061}$ & $\mathbf{0.702 \pm 0.103}$ & $\mathbf{0.943 \pm 0.064}$ \\
\hline
\end{tabular}
\vspace{-0.5em}
\end{table}

\begin{table}[t]
\centering
\caption{Venous $\to$ Home CGM: Insulin Resistance.}
\label{tab:venous_to_home_cgm_ir}
\footnotesize
\begin{tabular}{lccc}
\hline
\textbf{Method} & \textbf{AUROC} & \textbf{F1-score} & \textbf{PRAUC} \\
\hline
PCA & $0.827 \pm 0.146$ & $0.659 \pm 0.111$ & $0.847 \pm 0.125$ \\
\textsc{GluFormer} & $0.530 \pm 0.196$ & $0.537 \pm 0.140$ & $0.555 \pm 0.162$ \\
TS2Vec & $0.713 \pm 0.181$ & $0.588 \pm 0.158$ & $0.720 \pm 0.142$ \\
$\text{MOMENT}_{\text{Small}}$ & $0.830 \pm 0.192$ & $0.667 \pm 0.188$ & $0.852 \pm 0.141$ \\
$\text{MOMENT}_{\text{Large}}$ & $0.713 \pm 0.216$ & $0.556 \pm 0.233$ & $0.750 \pm 0.180$ \\
Mantis & $0.766 \pm 0.146$ & $0.649 \pm 0.178$ & $0.809 \pm 0.109$ \\
\hline
CGM-JEPA & $\mathbf{0.866 \pm 0.114}$ & \underline{$0.671 \pm 0.094$} & \underline{$0.892 \pm 0.086$} \\
X-CGM-JEPA & \underline{$0.865 \pm 0.115$} & $\mathbf{0.689 \pm 0.092}$ & $\mathbf{0.892 \pm 0.087}$ \\
\hline
\end{tabular}
\vspace{-0.5em}
\end{table}

\paragraph{$\beta$-cell Dysfunction.}
Under venous-to-CGM transfer (Table~\ref{tab:venous_to_home_cgm_beta}), \textsc{X-CGM-JEPA} is best on all three metrics and \textsc{CGM-JEPA} is second, forming a tight pair (X-vanilla $\Delta$AUROC $+0.003$, $\Delta$F1 $+0.005$). Compared to the strongest baseline (PCA), \textsc{X-CGM-JEPA} improves AUROC by \textbf{+2.2}~pp, F1 by \textbf{+1.1}~pp, and PRAUC by \textbf{+3.2}~pp. The more informative pattern is variance, not the mean: \textsc{X-CGM-JEPA} attains AUROC std $0.061$, against $0.074$ for PCA, $0.129$ for \textsc{GluFormer}, and $0.229$ for $\textsc{MOMENT}_{\text{Large}}$. Transfer is precisely the regime where high-variance behavior would most concern a deployer, and it is here that the JEPA family is most stable.

\paragraph{Insulin Resistance.}
For IR transfer (Table~\ref{tab:venous_to_home_cgm_ir}), \textsc{CGM-JEPA} achieves the best AUROC ($0.866$) and \textsc{X-CGM-JEPA} the best F1 ($0.689$) and PRAUC ($0.892$, tied). Against the strongest baseline ($\textsc{MOMENT}_{\text{Small}}$ at $0.830$), the JEPA family improves AUROC by \textbf{+3.6}~pp, F1 by \textbf{+2.2}~pp, and PRAUC by \textbf{+4.0}~pp. Two baselines collapse under IR transfer: \textsc{GluFormer} drops to $0.530$ (near-random) and $\textsc{MOMENT}_{\text{Large}}$ to $0.713$, both with AUROC std above $0.19$. The JEPA family holds AUROC std at $0.114$--$0.115$, the lowest among all methods in this cell, indicating that abstraction-first pretraining delivers stable transfer where reconstruction-based and broad time-series baselines do not.

Across both endpoints, transfer is the regime in which the JEPA family's advantage is largest in absolute terms and most uniform across metrics: \textsc{X-CGM-JEPA} or \textsc{CGM-JEPA} ranks first on every metric in both endpoints, with the cross-view variant contributing consistent F1 gains over \textsc{CGM-JEPA} under modality shift, consistent with the additive cross-view design.

\subsection{Cohort Generalization}
\label{subsec:venous}

We finally evaluate the cohort-generalization regime: encoders are pretrained as before, and downstream classifiers are trained on the Initial cohort venous data and tested on the Validation cohort venous data. Unlike the previous two regimes, this is a capability check rather than a deployment scenario, since population-scale screening cannot rely on venous OGTT at inference. The setting is informative for two reasons. First, the venous modality is the gold-standard supervision source, so strong cohort generalization here is necessary for downstream transfer to be meaningful. Second, venous sampling is much sparser than continuous CGM (CGM samples at a $5$-minute interval, while venous OGTT yields only a few discrete timepoints per session), making this the regime in which the cross-view design's promise of additive distributional structure from a complementary view is most directly testable.

\begin{table}[t]
\centering
\caption{In-Domain Venous (Cohort Generalization): $\beta$-cell Dysfunction.}
\label{tab:venous_in_domain_beta}
\footnotesize
\begin{tabular}{lccc}
\hline
\textbf{Method} & \textbf{AUROC} & \textbf{F1-score} & \textbf{PRAUC} \\
\hline
PCA & $0.790 \pm 0.109$ & $0.618 \pm 0.096$ & $0.784 \pm 0.101$ \\
\textsc{GluFormer} & $0.742 \pm 0.182$ & $0.598 \pm 0.121$ & $0.703 \pm 0.197$ \\
TS2Vec & $0.735 \pm 0.180$ & $0.555 \pm 0.192$ & $0.716 \pm 0.171$ \\
$\text{MOMENT}_{\text{Small}}$ & $0.712 \pm 0.029$ & $0.574 \pm 0.047$ & $0.668 \pm 0.047$ \\
$\text{MOMENT}_{\text{Large}}$ & $0.683 \pm 0.055$ & $0.570 \pm 0.061$ & $0.604 \pm 0.091$ \\
Mantis & $0.674 \pm 0.115$ & $0.560 \pm 0.066$ & $0.634 \pm 0.141$ \\
\hline
CGM-JEPA & \underline{$0.845 \pm 0.112$} & \underline{$0.645 \pm 0.084$} & $\mathbf{0.821 \pm 0.113}$ \\
X-CGM-JEPA & $\mathbf{0.855 \pm 0.064}$ & $\mathbf{0.664 \pm 0.061}$ & \underline{$0.815 \pm 0.080$} \\
\hline
\end{tabular}
\vspace{-0.5em}
\end{table}

\begin{table}[t]
\centering
\caption{In-Domain Venous (Cohort Generalization): Insulin Resistance.}
\label{tab:venous_in_domain_ir}
\footnotesize
\begin{tabular}{lccc}
\hline
\textbf{Method} & \textbf{AUROC} & \textbf{F1-score} & \textbf{PRAUC} \\
\hline
PCA & $0.744 \pm 0.014$ & $0.610 \pm 0.042$ & $0.770 \pm 0.032$ \\
\textsc{GluFormer} & $0.709 \pm 0.054$ & $0.580 \pm 0.053$ & $0.758 \pm 0.064$ \\
TS2Vec & $0.733 \pm 0.043$ & $0.621 \pm 0.040$ & $0.741 \pm 0.058$ \\
$\text{MOMENT}_{\text{Small}}$ & $0.674 \pm 0.049$ & $0.576 \pm 0.055$ & $0.671 \pm 0.075$ \\
$\text{MOMENT}_{\text{Large}}$ & $0.630 \pm 0.063$ & $0.611 \pm 0.056$ & $0.579 \pm 0.070$ \\
Mantis & $0.709 \pm 0.061$ & $0.582 \pm 0.076$ & $0.687 \pm 0.091$ \\
\hline
CGM-JEPA & $\mathbf{0.801 \pm 0.009}$ & \underline{$0.631 \pm 0.035$} & $\mathbf{0.824 \pm 0.008}$ \\
X-CGM-JEPA & \underline{$0.800 \pm 0.014$} & $\mathbf{0.653 \pm 0.046}$ & \underline{$0.822 \pm 0.020$} \\
\hline
\end{tabular}
\vspace{-0.5em}
\end{table}

\paragraph{$\beta$-cell Dysfunction.}
On the cohort-generalization $\beta$-cell task (Table~\ref{tab:venous_in_domain_beta}), \textsc{X-CGM-JEPA} achieves the best AUROC ($0.855$) and F1 ($0.664$), with \textsc{CGM-JEPA} second on both and best on PRAUC. Compared to the strongest baseline (PCA at $0.790$), the JEPA family improves AUROC by \textbf{+6.5}~pp, F1 by \textbf{+4.6}~pp, and PRAUC by \textbf{+3.7}~pp, the largest absolute downstream gains in the paper. The cross-view contribution is also most visible in this regime on a metric beyond F1: \textsc{X-CGM-JEPA} reduces AUROC standard deviation from $0.112$ (\textsc{CGM-JEPA}) to $0.064$, a $43\%$ relative reduction in fold-to-fold variance from the same encoder architecture under the same protocol. This is consistent with the additive cross-view design: when the temporal view is sparse, the distributional view contributes complementary structure that stabilizes the representation across cross-validation splits.

\paragraph{Insulin Resistance.}
For IR cohort generalization (Table~\ref{tab:venous_in_domain_ir}), \textsc{CGM-JEPA} and \textsc{X-CGM-JEPA} are tied on AUROC ($0.801$ vs $0.800$) and PRAUC ($0.824$ vs $0.822$), with \textsc{X-CGM-JEPA} clearly ahead on F1 ($0.653$ vs $0.631$, $+2.2$~pp). Against the strongest baseline (PCA at $0.744$), the JEPA family improves AUROC by \textbf{+5.7}~pp, F1 by \textbf{+3.2}~pp (over TS2Vec), and PRAUC by \textbf{+5.4}~pp. As in the previous two regimes, F1 is where \textsc{X-CGM-JEPA} contributes its most consistent gain over \textsc{CGM-JEPA}, accumulating to a within-family pattern of $+0.005$ ($\beta$-cell home), $+0.018$ (IR transfer), $+0.019$ ($\beta$-cell venous), and $+0.022$ (IR venous) across the four cells where the cross-view variant is not directly tied or ahead on AUROC.

The cohort-generalization regime delivers the largest AUROC gains in the paper but, equally importantly, it is where the cross-view design's distinctive contribution becomes quantitatively visible beyond F1: a halving of fold-to-fold variance on $\beta$-cell, and the most concentrated F1 gain pattern across the family. We trace these effects to the representation level in Section~\ref{subsec:representation-analysis}, where the additive distributional view leaves its strongest fingerprint on label-aware clustering structure.

\subsection{Representation Quality Analysis}
\label{subsec:representation-analysis}

To complement downstream classification, we examine the intrinsic geometry of learned embeddings using three families of unsupervised metrics: clustering quality (Silhouette, Calinski--Harabasz, Davies--Bouldin), distance-based structure (Between/Within ratio, Intra/Inter-cluster distance), and label-aware clustering agreement (Adjusted Rand Index, Normalized Mutual Information). The first two characterize how compact and well-separated the embeddings are without reference to outcome labels; the third tests whether the unsupervised cluster structure aligns with the clinical labels themselves. We compute all metrics on representations pooled across both outcomes (insulin resistance and $\beta$-cell dysfunction), reported separately by cohort and modality.

\begin{table*}[t]
\centering
\caption{Core clustering-based representation metrics. Higher is better for Silhouette (Sil) and Calinski--Harabasz (CH), while lower is better for Davies--Bouldin (DB). \emph{Bold/underline: best/second-best per dataset--modality.}}
\label{tab:core_metrics_by_modality}
\footnotesize
\setlength{\tabcolsep}{6pt}
\resizebox{\textwidth}{!}{%
\begin{tabular}{lccc lccc lccc}
\toprule
\multicolumn{4}{c}{\textbf{Initial cohort -- Venous}} &
\multicolumn{4}{c}{\textbf{Validation cohort -- CGM}} &
\multicolumn{4}{c}{\textbf{Validation cohort -- Venous}} \\
\cmidrule(lr){1-4}\cmidrule(lr){5-8}\cmidrule(lr){9-12}
Model & Sil $\uparrow$ & CH $\uparrow$ & DB $\downarrow$ &
Model & Sil $\uparrow$ & CH $\uparrow$ & DB $\downarrow$ &
Model & Sil $\uparrow$ & CH $\uparrow$ & DB $\downarrow$ \\
\midrule
PCA       & 0.1649 & 9.1153  & 1.5843 &
PCA       & 0.3022 & 10.1571 & 1.1455 &
PCA       & 0.1818 & 6.4728  & 1.3882 \\
GluFormer & 0.0652 & 3.5545  & 2.9108 &
GluFormer & 0.1073 & 3.5875  & 1.8875 &
GluFormer & 0.1022 & 3.1786  & 2.1317 \\
TS2Vec    & 0.0395 & 2.2737  & 3.7784 &
TS2Vec    & 0.1443 & 3.5651  & 1.9525 &
TS2Vec    & 0.0875 & 2.3896  & 2.4046 \\
Mantis    & 0.0262 & 1.7092  & 3.7788 &
Mantis    & 0.0723 & 2.2533  & 2.4958 &
Mantis    & 0.0520 & 1.7887  & 2.8017 \\
MOMENT    & 0.1013 & 3.0752  & 2.6811 &
MOMENT    & 0.0826 & 3.3272  & 1.8629 &
MOMENT    & 0.0423 & 1.9287  & 2.6529 \\
\cmidrule(lr){1-4}\cmidrule(lr){5-8}\cmidrule(lr){9-12}
CGM-JEPA  & \underline{0.1859} & \underline{11.4096} & \underline{1.4805} &
CGM-JEPA  & \textbf{0.3756}    & \underline{15.4511} & \underline{0.9123} &
CGM-JEPA  & \textbf{0.2220}    & \textbf{8.3631}     & \textbf{1.2411} \\
X-CGM-JEPA& \textbf{0.1940}    & \textbf{11.8594}    & \textbf{1.4610} &
X-CGM-JEPA& \underline{0.3636} & \textbf{15.4819}    & \textbf{0.9033} &
X-CGM-JEPA& \underline{0.1991} & \underline{8.0749}  & \underline{1.2372} \\
\bottomrule
\end{tabular}%
}
\vspace{-0.5em}
\end{table*}

\begin{table*}[t]
\centering
\caption{Distance-based representation structure metrics. Higher is better for Between/Within (B/W) ratio and Inter-cluster distance, while lower is better for Intra-cluster distance. \emph{Bold/underline: best/second-best per dataset--modality.}}
\label{tab:distance_metrics_by_modality}
\footnotesize
\setlength{\tabcolsep}{6pt}
\resizebox{\textwidth}{!}{%
\begin{tabular}{lccc lccc lccc}
\toprule
\multicolumn{4}{c}{\textbf{Initial cohort -- Venous}} &
\multicolumn{4}{c}{\textbf{Validation cohort -- CGM}} &
\multicolumn{4}{c}{\textbf{Validation cohort -- Venous}} \\
\cmidrule(lr){1-4}\cmidrule(lr){5-8}\cmidrule(lr){9-12}
Model & B/W $\uparrow$ & Intra $\downarrow$ & Inter $\uparrow$ &
Model & B/W $\uparrow$ & Intra $\downarrow$ & Inter $\uparrow$ &
Model & B/W $\uparrow$ & Intra $\downarrow$ & Inter $\uparrow$ \\
\midrule
PCA       & 0.3717 & 3.4512  & 4.5982  &
PCA       & 0.7735 & 3.6451  & 6.4387  &
PCA       & 0.4933 & 3.9220  & 5.7331  \\
GluFormer & 0.1448 & 3.1492  & 2.3278  &
GluFormer & 0.2697 & 2.4202  & 2.5665  &
GluFormer & 0.2432 & 2.9239  & 2.8285  \\
TS2Vec    & 0.0925 & 1.9931 & 1.1624 &
TS2Vec    & 0.2690 & 1.9297 & 1.9826 &
TS2Vec    & 0.1805 & 1.9579 & 1.6346 \\
Mantis    & 0.0705 & 23.3958 & 12.4460 &
Mantis    & 0.1706 & 23.5024 & 18.9659 &
Mantis    & 0.1355 & 22.8127 & 16.4026 \\
MOMENT    & 0.1270 & 0.6956 & 0.5161 &
MOMENT    & 0.2482 & 0.6301 & 0.6869 &
MOMENT    & 0.1442 & 0.6397 & 0.5125 \\
\cmidrule(lr){1-4}\cmidrule(lr){5-8}\cmidrule(lr){9-12}
CGM-JEPA  & \underline{0.4640} & 2.8409 & 4.1533 &
CGM-JEPA  & \underline{1.1825} & 2.6162 & 5.8530 &
CGM-JEPA  & \textbf{0.6406}    & 3.0579 & 5.0624 \\
X-CGM-JEPA& \textbf{0.4819}    & 2.5221 & 3.7567 &
X-CGM-JEPA& \textbf{1.1845}    & 2.3348 & 5.2742 &
X-CGM-JEPA& \underline{0.6185} & 2.7167 & 4.5175 \\
\bottomrule
\end{tabular}%
}
\vspace{-0.5em}
\end{table*}

\begin{table*}[t]
\centering
\caption{Label-agreement clustering metrics (KMeans cluster assignments vs.\ true metabolic labels). Higher is better for both Adjusted Rand Index (ARI) and Normalized Mutual Information (NMI). \emph{Bold/underline: best/second-best per dataset--modality.} Ties between CGM-JEPA, X-CGM-JEPA, and PCA on the validation cohort arise because the 2-cluster KMeans partition happens to coincide across these well-structured embeddings at $n \!\approx\! 17$.}
\label{tab:label_agreement_by_modality}
\footnotesize
\setlength{\tabcolsep}{6pt}
\resizebox{\textwidth}{!}{%
\begin{tabular}{lcc lcc lcc}
\toprule
\multicolumn{3}{c}{\textbf{Initial cohort -- Venous}} &
\multicolumn{3}{c}{\textbf{Validation cohort -- CGM}} &
\multicolumn{3}{c}{\textbf{Validation cohort -- Venous}} \\
\cmidrule(lr){1-3}\cmidrule(lr){4-6}\cmidrule(lr){7-9}
Model & ARI $\uparrow$ & NMI $\uparrow$ &
Model & ARI $\uparrow$ & NMI $\uparrow$ &
Model & ARI $\uparrow$ & NMI $\uparrow$ \\
\midrule
PCA       & \underline{0.2246} & \underline{0.2013} &
PCA       & \textbf{0.4707}    & \textbf{0.3853} &
PCA       & \textbf{0.3091}    & \underline{0.2506} \\
GluFormer & 0.1567            & 0.1414             &
GluFormer & $-0.0441$         & 0.1159             &
GluFormer & $-0.0441$         & 0.1159 \\
TS2Vec    & 0.1562            & 0.1421             &
TS2Vec    & 0.3088            & 0.2605             &
TS2Vec    & \underline{0.3088} & \textbf{0.2605} \\
Mantis    & $-0.0262$         & 0.0116             &
Mantis    & 0.1765            & 0.1463             &
Mantis    & 0.0736            & 0.1380 \\
MOMENT    & 0.0956            & 0.1671             &
MOMENT    & 0.0071            & 0.2009             &
MOMENT    & $-0.0515$         & 0.0352 \\
\cmidrule(lr){1-3}\cmidrule(lr){4-6}\cmidrule(lr){7-9}
CGM-JEPA  & 0.2079            & 0.1773             &
CGM-JEPA  & \textbf{0.4707}   & \textbf{0.3853}    &
CGM-JEPA  & \textbf{0.3091}   & \underline{0.2506} \\
X-CGM-JEPA& \textbf{0.2881}   & \textbf{0.2486}    &
X-CGM-JEPA& \textbf{0.4707}   & \textbf{0.3853}    &
X-CGM-JEPA& \textbf{0.3091}   & \underline{0.2506} \\
\bottomrule
\end{tabular}%
}
\vspace{-0.5em}
\end{table*}

\paragraph{Clustering and Distance Structure.}
Tables~\ref{tab:core_metrics_by_modality} and~\ref{tab:distance_metrics_by_modality} show the JEPA family delivers the strongest geometric structure across all three cohort--modality blocks, with no block in which a baseline outranks both \textsc{CGM-JEPA} and \textsc{X-CGM-JEPA}. On the Initial cohort venous block, \textsc{X-CGM-JEPA} is best on every geometric metric (Sil $0.194$, CH $11.86$, DB $1.46$, B/W $0.48$), with \textsc{CGM-JEPA} second across all of them. On the Validation cohort CGM block (the deployment-relevant modality), the two variants split the wins: \textsc{CGM-JEPA} attains the best Silhouette while \textsc{X-CGM-JEPA} attains the best CH and DB, and both lift the B/W ratio by over $50\%$ relative to PCA ($1.18$ vs.\ $0.77$). On the Validation cohort venous block, \textsc{CGM-JEPA} is best on all three clustering metrics and on B/W. The geometric advantage is therefore not specific to any one regime: predictive abstraction yields embeddings that are both compact and well-separated regardless of cohort or modality.

\paragraph{Label-aware Clustering Agreement.}
Geometric metrics measure how cluster-like the embedding is; they do not measure whether the clusters correspond to clinical labels. We therefore add a label-aware analysis (Table~\ref{tab:label_agreement_by_modality}): we run a 2-cluster KMeans on each embedding and measure agreement with ground-truth metabolic labels via ARI and NMI. The Initial cohort venous block reveals the clearest cross-view signal in the paper: \textsc{X-CGM-JEPA} achieves ARI $0.288$ and NMI $0.249$, against \textsc{CGM-JEPA} at $0.208$ and $0.177$, a relative improvement of $+39\%$ ARI and $+40\%$ NMI from the cross-view objective alone. PCA reaches ARI $0.225$, also below \textsc{X-CGM-JEPA}. On the Validation cohort blocks, the three top embeddings (PCA, \textsc{CGM-JEPA}, \textsc{X-CGM-JEPA}) coincide on the same KMeans partition due to the small subject count and well-structured 2-cluster geometry, producing identical ARI/NMI; this is a measurement-saturation artifact of small $n$ rather than a genuine tie of representational quality, which the geometric metrics already differentiate. The location of the label-aware signal is itself informative: the Initial cohort venous block is precisely the regime in which the temporal view is sparsest, and it is exactly here that the auxiliary distributional view contributes label-aligned structure that the temporal-only encoder does not produce on its own.

\paragraph{Synthesis.}
The three families of metrics tell a consistent two-part story. The JEPA family produces geometrically stronger embeddings than every baseline across every cohort--modality block, supporting the consistency claim at the representation level. Within the JEPA family, the cross-view objective leaves its quantitatively largest fingerprint on label-aware clustering on the sparse Initial-cohort venous data, supporting the additive-abstraction claim: when the temporal view is data-thin, a complementary distributional view contributes structure that aligns the embedding more closely with clinical labels. These representation-level findings are consistent with the downstream patterns observed in Sections~\ref{subsec:home-cgm}--\ref{subsec:venous}: the JEPA family delivers consistency across deployment regimes, while the cross-view extension contributes its most distinctive value where the temporal view is least informative.

\subsection{Where Discriminative Signal Concentrates: Per-Patch Label Divergence}
\label{subsec:patch-divergence}

To localize \emph{when} during the OGTT day the learned representations carry class-discriminative information, we compute per-patch label divergence: for each of the four post-meal temporal patches (P0--P3), we calculate the cosine distance between class-conditional mean embeddings on the test set. Higher divergence at a patch means the encoder distinguishes the two classes more sharply at that time window. Tables~\ref{tab:patch_div_beta} and~\ref{tab:patch_div_ir} report patch-wise divergence for the two endpoints.

\begin{table}[t]
\centering
\caption{Per-patch label divergence: $\beta$-cell Dysfunction. Higher = more class-discriminative.}
\label{tab:patch_div_beta}
\footnotesize
\setlength{\tabcolsep}{4pt}
\begin{tabular}{lcccc}
\hline
\textbf{Encoder} & \textbf{P0} & \textbf{P1} & \textbf{P2} & \textbf{P3}$^\dagger$ \\
 & \scriptsize $-10$--45\,m & \scriptsize 50--105\,m & \scriptsize 110--165\,m & \scriptsize 170--225\,m \\
\hline
CGM-JEPA   & $0.009$ & $0.159$ & $\mathbf{0.161}$ & $0.014$ \\
X-CGM-JEPA & $0.004$ & $0.130$ & $\mathbf{0.161}$ & $0.008$ \\
\hline
\end{tabular}\\[0.3em]
{\scriptsize $^\dagger$P3 has only 3 real venous observations ($t=170,175,180$); the remaining steps are flat-padded at the final glucose value, so P3 divergence is not directly comparable across patches.}
\vspace{-0.5em}
\end{table}

\begin{table}[t]
\centering
\caption{Per-patch label divergence: Insulin Resistance. Higher = more class-discriminative.}
\label{tab:patch_div_ir}
\footnotesize
\setlength{\tabcolsep}{4pt}
\begin{tabular}{lcccc}
\hline
\textbf{Encoder} & \textbf{P0} & \textbf{P1} & \textbf{P2} & \textbf{P3}$^\dagger$ \\
 & \scriptsize $-10$--45\,m & \scriptsize 50--105\,m & \scriptsize 110--165\,m & \scriptsize 170--225\,m \\
\hline
CGM-JEPA   & $0.140$ & $\mathbf{0.448}$ & $0.292$ & $0.023$ \\
X-CGM-JEPA & $0.109$ & $\mathbf{0.373}$ & $0.288$ & $0.012$ \\
\hline
\end{tabular}\\[0.3em]
{\scriptsize $^\dagger$Same padding caveat as Table~\ref{tab:patch_div_beta}.}
\vspace{-0.5em}
\end{table}

\paragraph{Endpoint-specific Temporal Localization.}
The two endpoints exhibit visibly different patch-divergence profiles. For insulin resistance, divergence peaks at \textbf{P1 ($50$--$105$\,min)}, the early post-load phase, with a secondary plateau at P2 ($110$--$165$\,min). For $\beta$-cell dysfunction, divergence peaks instead at \textbf{P2}, with P1 already substantially above the early P0 baseline. The pattern is consistent across both encoders. This timing difference is consistent with what is known about the two failure modes physiologically: peripheral glucose-clearance impairment (IR) becomes apparent immediately after the glucose load, whereas inadequate insulin secretion ($\beta$-cell dysfunction) becomes more apparent once the response should be fully mobilized. We stress that this is a representation-level observation rather than a direct mechanistic test, but the agreement with prior physiological understanding indicates that the encoders are extracting label-relevant signal from clinically interpretable time windows rather than from spurious correlates.

\paragraph{Cross-view Effect on Temporal Localization.}
Comparing the two encoders, \textsc{X-CGM-JEPA} consistently shows \emph{lower} peak per-patch divergence than \textsc{CGM-JEPA} (e.g., IR P1: $0.373$ vs.\ $0.448$). This is informative given that the two encoders achieve comparable downstream AUROC and that \textsc{X-CGM-JEPA} attains better F1 on the same task: lower per-patch divergence with equal or better thresholded performance suggests that the cross-view objective spreads class-discriminative signal more broadly across the day rather than concentrating it in a single patch. This pattern is consistent with the design intent of \textsc{X-CGM-JEPA}: by adding a distributional view that integrates evidence across the entire day, the cross-view objective discourages the encoder from leaning on any single time-window's signature.

\paragraph{Limitations.}
Per-patch divergence is a coarse, patch-aggregated statistic on a small test cohort and does not establish a causal mechanism. P3 divergence values (Tables~\ref{tab:patch_div_beta},~\ref{tab:patch_div_ir}) are particularly difficult to interpret due to the padding artifact noted in the footnote. We therefore present this analysis as suggestive rather than definitive evidence of the temporal physiological signature each endpoint elicits, and as a starting point for finer-grained interpretability work.

\subsection{Demographic Subgroup Redistribution}
\label{subsec:subgroup-redistribution}

Sections~\ref{subsec:home-cgm} and~\ref{subsec:transfer} established that \textsc{X-CGM-JEPA} matches \textsc{CGM-JEPA} on mean AUROC under deployment but contributes a consistent F1 advantage. We now examine where this F1 advantage comes from at the subgroup level. We stratify post-hoc test-set predictions in the two CGM deployment regimes (Home-CGM in-domain and Venous-to-Home-CGM transfer) by sex, age band, BMI category, and ethnicity, reporting AUROC for every subgroup with $n \geq 5$ subjects. The pattern is consistent across both regimes: the cross-view objective performs a worst-group-first redistribution, lifting subgroups where \textsc{CGM-JEPA} underperforms while leaving already-strong subgroups largely unchanged.

\begin{table}[t]
\centering
\caption{Venous-to-Home-CGM transfer: subgroup AUROCs, X-CGM-JEPA vs.\ CGM-JEPA. $\Delta$ = X-CGM-JEPA $-$ CGM-JEPA. All subgroups with $n \geq 5$ shown.}
\label{tab:redistribution_transfer}
\footnotesize
\setlength{\tabcolsep}{4pt}
\begin{tabular}{llcccr}
\hline
\textbf{Endpoint} & \textbf{Subgroup} & $n$ & \textbf{CGM-JEPA} & \textbf{X-CGM-JEPA} & $\Delta$ \\
\hline
\multicolumn{6}{l}{\textit{$\beta$-cell Dysfunction}} \\
$\beta$ & Ethn.\ Asian       &  5 & $0.739$ & $0.792$ & $\mathbf{+0.052}$ \\
$\beta$ & Sex = F            & 10 & $0.761$ & $0.777$ & $+0.016$ \\
$\beta$ & Age 50--59         & 12 & $0.882$ & $0.895$ & $+0.013$ \\
$\beta$ & BMI 18.5--24.9     &  6 & $0.962$ & $0.972$ & $+0.010$ \\
$\beta$ & Sex = M            &  7 & $0.979$ & $0.978$ & $-0.001$ \\
$\beta$ & Ethn.\ Caucasian   & 12 & $0.985$ & $0.976$ & $-0.009$ \\
$\beta$ & BMI 25--29.9       &  6 & $0.999$ & $0.984$ & $-0.016$ \\
\hline
\multicolumn{6}{l}{\textit{Insulin Resistance}} \\
IR & Ethn.\ Asian         &  5 & $0.669$ & $0.723$ & $\mathbf{+0.054}$ \\
IR & Sex = F              & 10 & $0.615$ & $0.638$ & $+0.023$ \\
IR & BMI 25--29.9         &  6 & $0.966$ & $0.980$ & $+0.014$ \\
IR & Ethn.\ Caucasian     & 12 & $0.753$ & $0.762$ & $+0.009$ \\
IR & Age 50--59           & 12 & $0.753$ & $0.751$ & $-0.002$ \\
IR & Sex = M              &  7 & $0.914$ & $0.906$ & $-0.009$ \\
IR & BMI 18.5--24.9       &  6 & $0.900$ & $0.874$ & $-0.025$ \\
\hline
\end{tabular}
\vspace{-0.5em}
\end{table}

\begin{table}[t]
\centering
\caption{Home-CGM in-domain: subgroup AUROCs, X-CGM-JEPA vs.\ CGM-JEPA. All subgroups with $n \geq 5$ shown.}
\label{tab:redistribution_homecgm}
\footnotesize
\setlength{\tabcolsep}{4pt}
\begin{tabular}{llcccr}
\hline
\textbf{Endpoint} & \textbf{Subgroup} & $n$ & \textbf{CGM-JEPA} & \textbf{X-CGM-JEPA} & $\Delta$ \\
\hline
\multicolumn{6}{l}{\textit{Insulin Resistance}} \\
IR & Sex = F           & 10 & $0.529$ & $0.570$ & $\mathbf{+0.040}$ \\
IR & BMI 25--29.9      &  6 & $0.954$ & $0.985$ & $+0.031$ \\
IR & Age 50--59        & 12 & $0.747$ & $0.770$ & $+0.023$ \\
IR & Ethn.\ Caucasian  & 12 & $0.752$ & $0.769$ & $+0.016$ \\
IR & Ethn.\ Asian      &  5 & $0.716$ & $0.733$ & $+0.016$ \\
IR & Sex = M           &  7 & $0.973$ & $0.971$ & $-0.002$ \\
IR & BMI 18.5--24.9    &  6 & $0.973$ & $0.967$ & $-0.006$ \\
\hline
\multicolumn{6}{l}{\textit{$\beta$-cell Dysfunction (near-saturated)}} \\
$\beta$ & Sex = F                 & 10 & $0.709$ & $0.714$ & $+0.005$ \\
$\beta$ & Age 50--59              & 12 & $0.883$ & $0.888$ & $+0.005$ \\
$\beta$ & Ethn.\ Asian            &  5 & $0.783$ & $0.787$ & $+0.005$ \\
$\beta$ & Sex = M                 &  7 & $0.999$ & $0.999$ & $\phantom{-}0.000$ \\
$\beta$ & BMI 18.5--24.9          &  6 & $1.000$ & $1.000$ & $\phantom{-}0.000$ \\
$\beta$ & BMI 25--29.9            &  6 & $1.000$ & $1.000$ & $\phantom{-}0.000$ \\
$\beta$ & Ethn.\ Caucasian        & 12 & $1.000$ & $0.995$ & $-0.005$ \\
\hline
\end{tabular}
\vspace{-0.5em}
\end{table}

\paragraph{Cross-modality Transfer.}
Under transfer (Table~\ref{tab:redistribution_transfer}), the largest \textsc{X-CGM-JEPA} gains concentrate on the subgroups where \textsc{CGM-JEPA} is weakest. On both endpoints, the Asian-ethnicity subgroup, which has the lowest \textsc{CGM-JEPA} AUROC ($0.739$ on $\beta$-cell, $0.669$ on IR), receives the largest boost ($+0.052$ and $+0.054$ respectively). The female subgroup also improves on both endpoints ($+0.016$ on $\beta$-cell, $+0.023$ on IR), again starting from below the model's average performance. Offsetting decrements appear on already-strong majority subgroups (Caucasian-$\beta$ $-0.009$, BMI 25--29.9 $\beta$ $-0.016$, BMI 18.5--24.9 IR $-0.025$), but in absolute terms these subgroups remain at AUROC $> 0.87$.

The aggregate effect is a compression of the subgroup performance distribution rather than a uniform lift. The maximum-to-minimum ethnicity AUROC gap shrinks by $25\%$ on $\beta$-cell ($0.246 \to 0.184$) and $54\%$ on IR ($0.084 \to 0.039$). Sex gaps shrink by $8\%$ ($\beta$-cell) and $10\%$ (IR). Mean AUROC across subgroups changes by less than $+0.01$ on both endpoints. The cross-view objective therefore does not raise the average; it compresses the worst-to-best spread.

\paragraph{In-domain Home CGM.}
The same redistribution pattern appears in the home in-domain regime (Table~\ref{tab:redistribution_homecgm}), most clearly on insulin resistance. The female subgroup, on which \textsc{CGM-JEPA} performs near-chance ($0.529$), receives the largest \textsc{X-CGM-JEPA} lift ($+0.040$) and crosses above $0.57$, while male performance, at ceiling for \textsc{CGM-JEPA} ($0.973$), remains essentially unchanged ($-0.002$). The IR sex gap shrinks from $0.444$ to $0.401$ (a $10\%$ reduction). Both ethnicity subgroups improve symmetrically ($+0.016$). The $\beta$-cell endpoint is near-saturated in the home regime ($\geq 0.99$ on five of seven subgroups for both methods), so all $\beta$-cell deltas in this table fall within run-to-run noise and we draw no inference from them.

\paragraph{Caveats.}
Per-subgroup sample sizes are small ($n = 5$ to $12$), and individual cell AUROCs should be read with appropriate caution; we report all subgroups meeting $n \geq 5$ to avoid selection. The \emph{pattern} is robust across $20$ iterations $\times$ $2$-fold cross-validation: across both regimes and both endpoints, the largest \textsc{X-CGM-JEPA} gains land on subgroups where \textsc{CGM-JEPA} is weakest, and the largest decrements land on subgroups already near ceiling. The redistribution is consistent with the additive-cross-view design: the Glucodensity view supplies complementary structure that disproportionately benefits subgroups whose temporal patterns alone are harder for the encoder to separate, providing a deployment-relevant fairness property that the AUROC means do not capture.

\section{Ablation Studies}

\paragraph{Effect of Label Availability.}

% I updated the plot with only linear probe (Logistic Regression). Previously there were 5 classification heads averaged
\begin{figure}[t]
    \centering
    \includegraphics[width=0.9\linewidth]{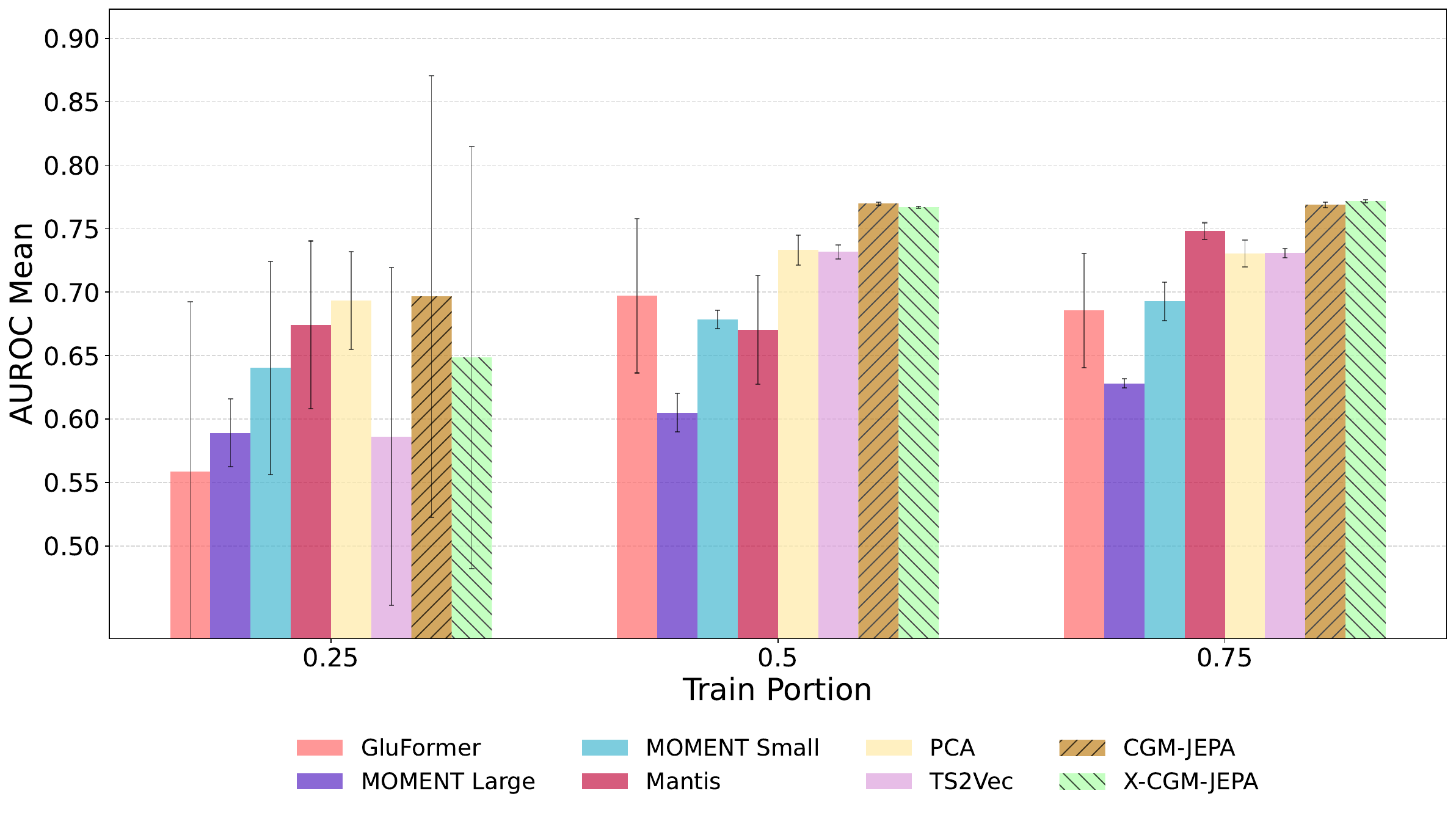}
    \caption{Comparison across different training portions.}
    \label{fig:label-scarcity}
    \vspace{-1em}
\end{figure}

\begin{figure}[!htbp]
    \centering
    \begin{subfigure}[t]{0.6\linewidth}
        \centering
        \includegraphics[width=\linewidth]{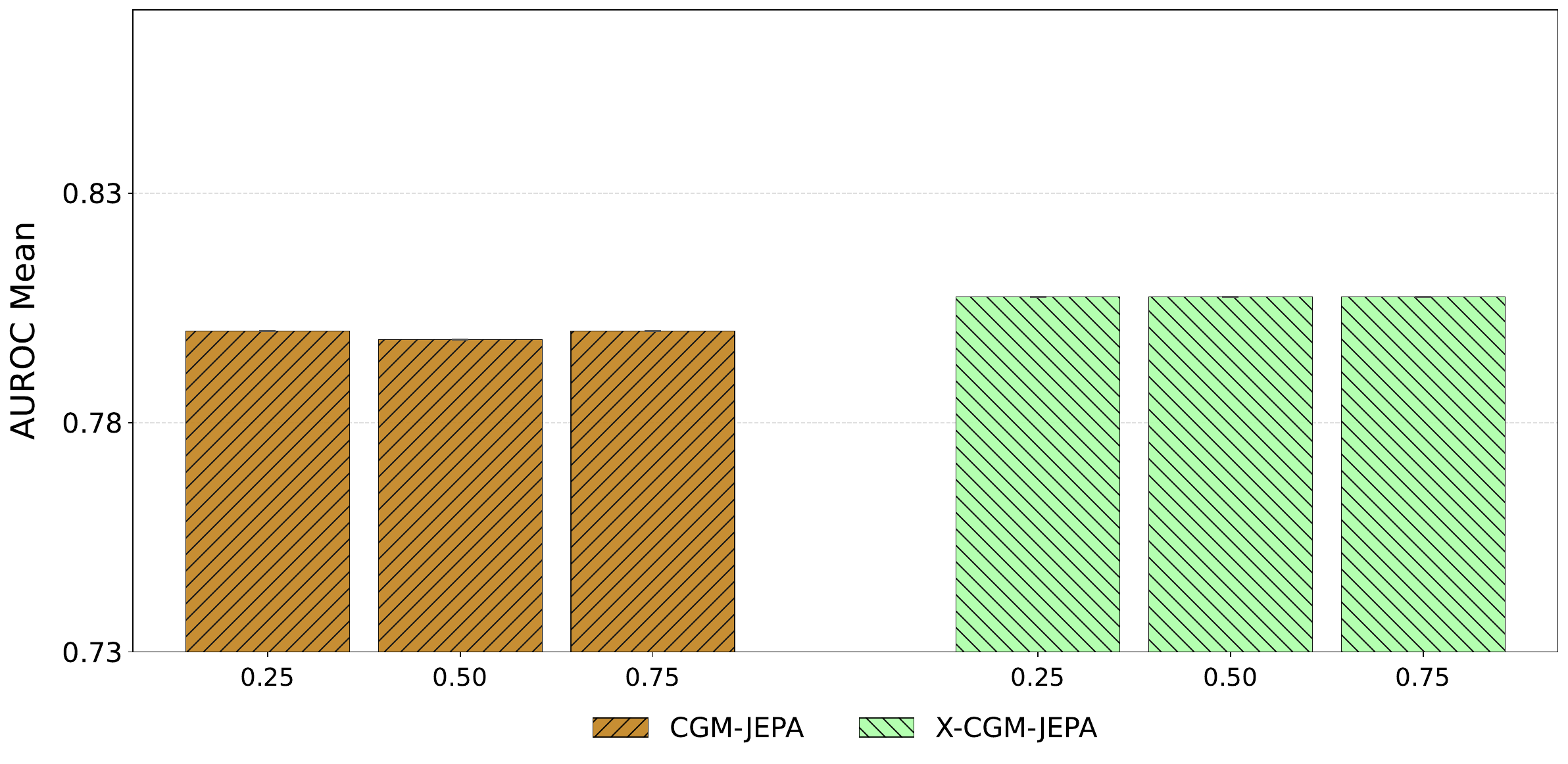}
        \caption{CGM-JEPA and X-CGM-JEPA: masking ratio sensitivity.}
        \label{fig:cross_jepa_mr}
    \end{subfigure}\hfill
    \begin{subfigure}[t]{0.38\linewidth}
        \centering
        \includegraphics[width=\linewidth]{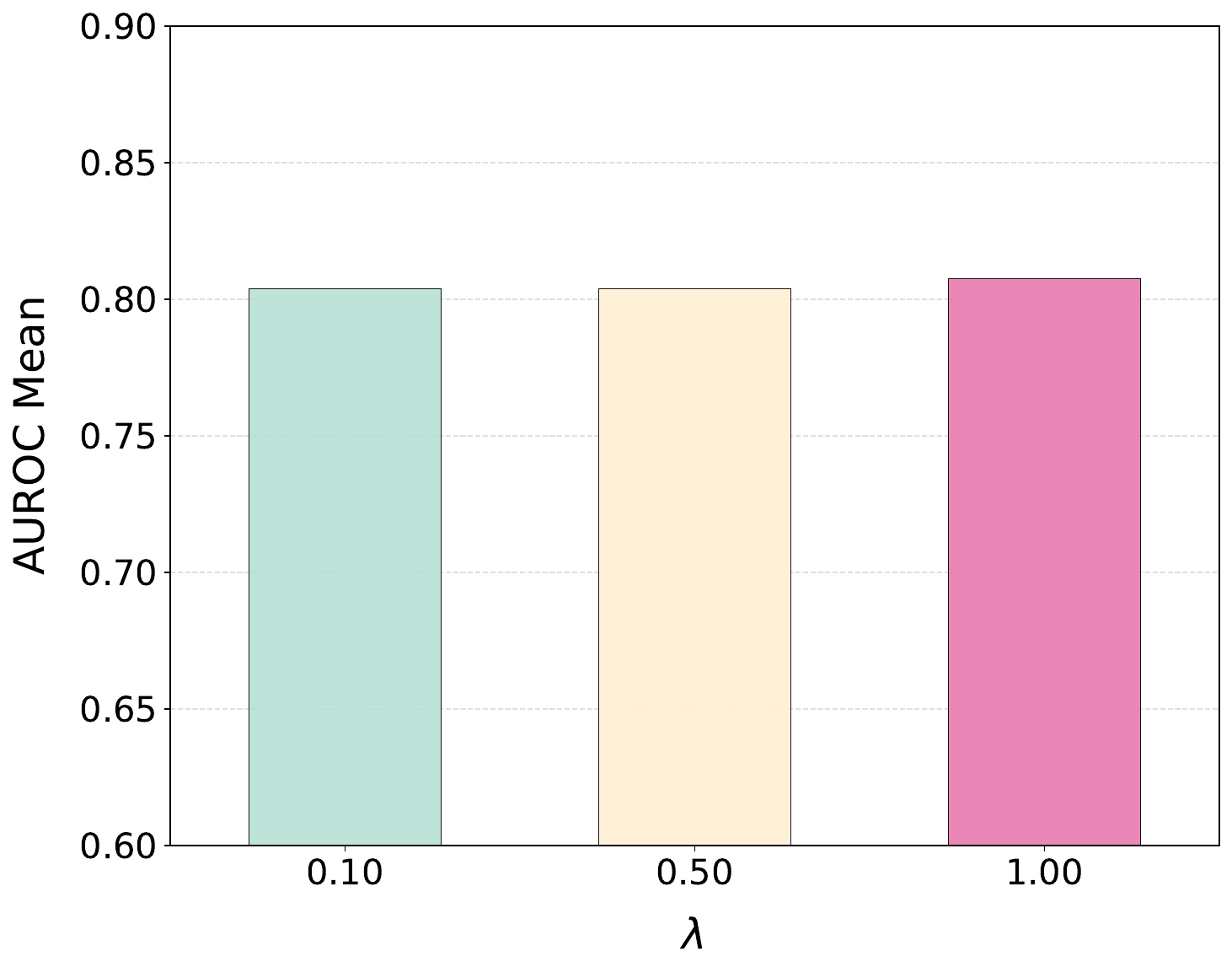}
        \caption{X-CGM-JEPA: cross-view weight $\lambda$ sensitivity.}
        \label{fig:cross_jepa_lambda_rocauc}
    \end{subfigure}
    \caption{Hyperparameter sensitivity of the JEPA family. (a) AUROC across masking ratios for both encoders. (b) \textsc{X-CGM-JEPA} sensitivity to the cross-view weight $\lambda$.}
    \label{fig:ablation_jepa}
    \vspace{-0.5em}
\end{figure}

Figure~\ref{fig:label-scarcity} reports AUROC and run-to-run standard deviation across labeled-data portions ($25\%$, $50\%$, $75\%$). The picture has two distinct regimes. At the $25\%$ portion (very few labeled subjects), all eight encoders sit within a noisy band: \textsc{CGM-JEPA} is nominally first ($0.697$) and PCA second ($0.693$), but the standard deviations are large for several methods (\textsc{CGM-JEPA} $0.174$, \textsc{X-CGM-JEPA} $0.166$, \textsc{TS2Vec} $0.133$, \textsc{GluFormer} $0.134$), reflecting how little training signal is available at this portion rather than a property of the encoders themselves. We do not draw rank-level conclusions in this regime.

Once labels reach $50\%$ or $75\%$, the picture stabilizes sharply and the JEPA family separates from the rest. At $50\%$: \textsc{CGM-JEPA} $0.770 \pm 0.001$ and \textsc{X-CGM-JEPA} $0.767 \pm 0.001$ are first and second, with the strongest baseline (PCA) at $0.733 \pm 0.012$. At $75\%$: \textsc{X-CGM-JEPA} $0.772 \pm 0.001$ and \textsc{CGM-JEPA} $0.769 \pm 0.002$ remain first and second; the next baseline (Mantis) reaches $0.748 \pm 0.007$. Two patterns are notable. First, the JEPA family consistently outperforms all baselines by $2$ to $4$ AUROC points whenever the data regime is informative enough to draw conclusions. Second, the JEPA family's run-to-run standard deviation is one to two orders of magnitude smaller than the strongest baselines at $50\%$ and $75\%$ (e.g., $0.0007$ vs.\ PCA $0.012$ at $50\%$), indicating that abstraction-first pretraining yields representations whose performance is highly reproducible across folds once supervision is sufficient to identify the embedding's discriminative structure.

\paragraph{Effect of Masking Ratio.}
We sweep the pretraining masking ratio over $\{25\%, 50\%, 75\%\}$ for both encoders (Figure~\ref{fig:ablation_jepa}a). AUROC is essentially flat: \textsc{CGM-JEPA} mean varies within $0.001$ across the three ratios, and \textsc{X-CGM-JEPA} produces identical mean AUROC ($0.805$) at all three. Run-to-run standard deviations are also small in absolute terms ($0.002$--$0.007$), and \textsc{X-CGM-JEPA} is approximately $2.7\times$ more stable than \textsc{CGM-JEPA} at every ratio ($0.0021$ vs.\ $0.0057$ on average). The flat behavior indicates that day-level CGM windows carry sufficient temporal redundancy for simple random masking to provide a reliable pretraining signal across a wide range, and the cross-view objective reduces variance further without altering the central tendency. We use $25\%$ masking in all main results on this basis.

\paragraph{Effect of Cross-view Weight $\lambda$.}
Figure~\ref{fig:ablation_jepa}b reports \textsc{X-CGM-JEPA} AUROC at $\lambda \in \{0.1, 0.5, 1.0\}$, averaged across masking ratios. Mean AUROC ranges over $[0.7730, 0.7745]$, a spread of $0.0016$, while the run-to-run standard deviation at each $\lambda$ is $0.043$--$0.046$, more than an order of magnitude larger than the mean spread. Differences across $\lambda$ are therefore well within fold-to-fold variability, and the three settings are not distinguishable in this dataset. The robustness is consistent with the additive-abstraction interpretation: the Glucodensity view contributes complementary structure rather than competing with the temporal objective, so the encoder is not destabilized by varying their relative weight. We fix $\lambda{=}1$ in all main results.
\section{Discussion}
\label{sec:discussion}

\paragraph{Limitations.} Our study leverages a clinically curated cohort that provides two forms of \emph{gold-standard} information: (i) \emph{gold-standard sample collection} via venous blood draws during an in-clinic OGTT, and (ii) \emph{gold-standard condition labels} (insulin resistance and $\beta$-cell dysfunction) derived from these assessments. Because OGTT-based protocols are costly and invasive, such high-fidelity labels are rarely available at scale; accordingly, our dataset serves as a high-quality benchmark for studying label-efficient CGM representation learning. Despite explicitly evaluating distribution shifts (e.g., venous-to-CGM and controlled-to-home), our experiments are still tied to a specific data-collection pipeline and sensor modality. Real-world deployment may introduce additional shifts (e.g., device calibration, missingness patterns) and subgroup-dependent performance. While we reduce overfitting and leakage via cohort hold-out evaluation, stratified cross-validation, and fold-wise normalization, limited subgroup metadata prevents comprehensive fairness analysis; larger multi-site, multi-device cohorts are needed for robust validation and bias assessment.

\paragraph{Conclusion.}
We studied metabolic dysfunction prediction from continuous glucose monitoring (CGM) under three clinically motivated settings: controlled venous evaluation with cohort generalization, venous-to-CGM transfer, and real-world home CGM evaluation. We introduced \textsc{CGM-JEPA}, a predictive self-supervised pretraining framework tailored to day-level CGM windows, and \textsc{X-CGM-JEPA}, which further regularizes CGM representations via an auxiliary cross-view Glucodensity prediction objective. Across experiments, predictive pretraining yields strong and label-efficient representations, while the cross-view extension provides the most consistent benefits under clinically relevant distribution shifts, particularly in venous-supervised deployment where inference relies on wearable CGM. In contrast, in-domain home evaluation exhibits limited headroom over strong baselines, suggesting that the main value of cross-view guidance lies in improving transferability rather than maximizing in-domain accuracy. 

Our results highlight the promise of predictive self-supervision for extracting clinically meaningful structure from consumer-grade CGM with scarce gold-standard labels. Future work will extend cross-view objectives to richer physiological signals and larger cohorts, and investigate uncertainty-aware deployment for individualized metabolic risk assessment.

\bibliography{ref}
\bibliographystyle{plain}

%%%%%%%%%%%%%%%%%%%%%%%%%%%%%%%%%%%%%%%%%%%%%%%%%%%%%%%%%%%%
%%% Methods (M.x numbering, ED figures/tables) %%%
\newpage
\renewcommand{\thesection}{M.\arabic{section}}
\renewcommand{\thefigure}{ED.\arabic{figure}}
\renewcommand{\thetable}{ED.\arabic{table}}
\renewcommand{\theequation}{M.\arabic{equation}}

\setcounter{section}{1}
\setcounter{figure}{0}
\setcounter{table}{0}
\setcounter{equation}{0}
\renewcommand{\theHsection}{M.\arabic{section}}
\renewcommand{\theHfigure}{ED.\arabic{figure}}
\renewcommand{\theHtable}{ED.\arabic{table}}

\noindent \textbf{\LARGE{Methods}}
\normalfont

\section{Related Work}
\label{related_work}

\textbf{CGM-Based Metabolic Subphenotype Prediction.}
Prior studies have demonstrated that continuous glucose monitoring (CGM) data can be leveraged to predict clinically meaningful metabolic subphenotypes using supervised machine learning. In particular, \cite{metwally2025prediction} shows that CGM-derived features can be used to infer insulin resistance and $\beta$-cell dysfunction when paired with gold-standard venous glucose measurements. While effective, such approaches rely on fully supervised learning and require access to high-quality metabolic labels, which are expensive and invasive to obtain. Consequently, their applicatbility is limited in large-scale or real-world settings where labeled data are scarce.

\textbf{Functional Representations of Glucose Dynamics.}
Beyond conventional summary statistics, functional representations of CGM have been proposed to better characterize glucose dynamics. Glucodensity-based methods model glucose trajectories through distributional profile, capturing the statistical structure of glucose fluctuations over time rather than point-wise values \cite{klonoff2025continuous}. Empirical evidence suggests that glucodensity functional profiles can outperform traditional CGM metrics in downstream characterization tasks \cite{matabuena2025glucodensity}. These findings highlight glucodensity as a complementary view of glucose regulation, motivating its use alongside raw CGM signals to capture physiologically relevant dynamics.

\textbf{Self-Supervised Learning for CGM Time Series.}
Self-supervised learning has recently emerged as a promising approach for learning CGM representations without extensive labeled annotations. \textsc{GluFormer}~\cite{lutsker2025gluformer} pretrains models with autoregressive and reconstruction-style objectives, emphasizing point-wise signal fidelity and capturing short-term glucose dynamics. In parallel, \textsc{CGMformer}~\cite{lu2025pretrained} leverages large-scale CGM pretraining to learn individualized glucose representations and demonstrates improved generalization across downstream tasks.

\textbf{Time Series Foundation Model.} Beyond CGM-specific SSL, recent time-series foundation models pretrained on large, multi-domain corpora aim to provide transferable representations with strong out-of-the-box performance on downstream tasks. Representative examples include MOMENT~\cite{goswami2024moment} and Mantis~\cite{feofanov2025mantislightweightcalibratedfoundation} for general representation learning and classification, as well as forecasting-oriented pretrained models such as TimesFM~\cite{10.5555/3692070.3692474} and Chronos~\cite{ansari2024chronos}. Related sensor-based language models have also been proposed for wearable and physiological data~\cite{li-etal-2025-sensorllm,zhang2025sensorlmlearninglanguagewearable,li2025zarazeroshotmotiontimeseries}. While these models benefit from scale and broad coverage, their pretraining data and objectives are not tailored to CGM's physiology-driven dynamics or to device-specific noise, missingness, and sampling irregularities.

\section{Methodology}
\label{methods}

\begin{figure*}[t]
    \centering
    \includegraphics[width=\textwidth]{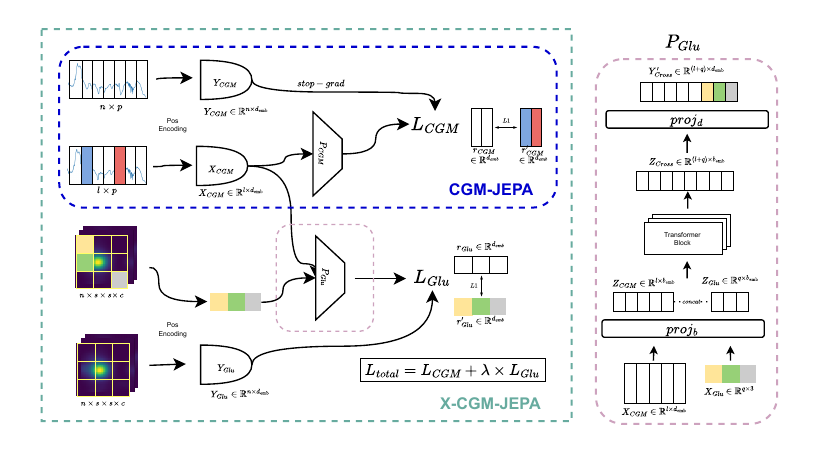}
    \caption{Overview of our predictive pretraining framework for CGM. \textsc{CGM-JEPA} learns representations by predicting masked CGM patch embeddings from visible context. \textsc{X-CGM-JEPA} extends \textsc{CGM-JEPA} with an auxiliary cross-view objective that predicts a masked Glucodensity embedding derived from the same day window.}
    \label{fig:cross-view-jepa-architecture}
    \vspace{-0.5em}
\end{figure*}

\subsection{Problem Setup}
For each subject $i$, we observe a CGM time series $\mathbf{X}_i=\{g_{i1},\ldots,g_{iT_i}\}$, where $g_{it}\in\mathbb{R}$ is the glucose value recorded at the $t$-th timestamp. A subset of subjects has gold-standard metabolic labels for two binary outcomes: insulin resistance and $\beta$-cell dysfunction. Our goal is to learn an encoder $f_\theta$ that maps a CGM window to a fixed-dimensional embedding $\mathbf{z}_i=f_\theta(\mathbf{X}_i)\in\mathbb{R}^d$, enabling a lightweight classifier $h_\phi$ to predict these outcomes with limited supervision.
Because labeled metabolic assessments are scarce, we learn $f_\theta$ via self-supervised pretraining on unlabeled CGM and evaluate representation quality via linear probing.

\subsection{CGM Windowing and Tokenization}
We segment each CGM stream into daily windows of length 288 (5-minute sampling). Each day is tokenized into $P=24$ non-overlapping hourly patches with patch size 12. Let $\mathbf{x}\in\mathbb{R}^{P\times 12}$ denote a tokenized day. This patchified representation standardizes temporal resolution and supports patch-level masking for predictive pretraining.

\subsection{Predictive Pretraining on CGM: \textsc{CGM-JEPA}}
\label{sec:cgm-jepa}
We first introduce \textsc{CGM-JEPA}, a JEPA-style predictive representation learner tailored to CGM. Rather than reconstructing raw glucose values, \textsc{CGM-JEPA} performs \emph{masked representation prediction}: it predicts the latent representations of masked patches from the visible context, encouraging the encoder to capture higher-level temporal structure.

\subsubsection{\textbf{Masked context--target construction.}}
Given a tokenized day $\mathbf{x}\in\mathbb{R}^{P\times 12}$, we sample a patch mask $\mathcal{M}\subseteq\{1,\dots,P\}$ and form a context view $\mathbf{x}^{(c)}$ by retaining only visible patches, while the target view corresponds to the masked patch indices $\mathcal{M}$. A context encoder $f_\theta$ encodes the visible context, and a target encoder $f_{\bar{\theta}}$ encodes the target view:
\[
\mathbf{z}^{(c)} = f_\theta(\mathbf{x}^{(c)}), 
\qquad 
\mathbf{z}^{(t)} = f_{\bar{\theta}}(\mathbf{x}),
\]
where $\mathbf{z}^{(t)}=\{\mathbf{z}^{(t)}_j\}_{j=1}^{P}$ provides per-patch target representations and only $\{\mathbf{z}^{(t)}_j\}_{j\in\mathcal{M}}$ are used as prediction targets. A predictor $p_\phi$ maps the context embedding to predicted representations for the masked indices:
\[
\hat{\mathbf{z}}^{(t)} = p_\phi(\mathbf{z}^{(c)}), \qquad 
\hat{\mathbf{z}}^{(t)}=\{\hat{\mathbf{z}}^{(t)}_j\}_{j\in\mathcal{M}}.
\]
We minimize an $\ell_1$ (MAE) regression loss between predicted and target representations over masked patches, stopping gradients on the target branch:
\[
\mathcal{L}_{\text{CGM}}
=
\frac{1}{|\mathcal{M}|}
\sum_{j\in\mathcal{M}}
\left\lVert
\hat{\mathbf{z}}^{(t)}_{j}
-
\operatorname{stopgrad}\!\left(\mathbf{z}^{(t)}_{j}\right)
\right\rVert_{1}.
\]
This objective yields a predictive SSL framework for CGM that can be trained on abundant unlabeled home CGM.

\subsection{Cross-view Regularization: \textsc{X-CGM-JEPA}}
\label{sec:cross-jepa}
While \textsc{CGM-JEPA} focuses on temporal structure in the 1D trajectory, CGM windows also exhibit clinically meaningful distributional patterns (e.g., variability and density over time). To capture complementary distributional structure, especially under modality or setting shifts, we propose \textsc{X-CGM-JEPA}, which augments \textsc{CGM-JEPA} with an auxiliary cross-view prediction objective based on a Glucodensity representation.

\subsubsection{\textbf{Glucodensity view.}}
From the same daily CGM window, we derive a Glucodensity representation $\mathbf{D}$ using our preprocessing pipeline. $\mathbf{D}$ is a deterministic transformation that summarizes distributional structure of glucose dynamics over the day. We patchify $\mathbf{D}$ into tokens and apply patch-level masking, denoting the masked Glucodensity tokens by $\mathbf{D}^{(t)}$. A Glucodensity encoder $g_\psi$ maps the masked Glucodensity tokens to an embedding:
\[
\mathbf{u} = g_\psi(\mathbf{D}^{(t)}).
\]

\subsubsection{\textbf{Asymmetric cross-view prediction.}}
Using the CGM context embedding $\mathbf{z}^{(c)}$, a cross-view predictor $q_\omega$ predicts the masked Glucodensity embedding:
\[
\hat{\mathbf{u}} = q_\omega(\mathbf{z}^{(c)}), 
\qquad
\mathcal{L}_{\text{GD}}
=
\left\lVert \hat{\mathbf{u}}-\mathbf{u}\right\rVert_{1}.
\]
Here, Glucodensity is used as an auxiliary \emph{target}, and masking is applied to mitigate shortcuts from full-window summaries. Unlike the CGM target branch, gradients are allowed to flow into $g_\psi$ so that the Glucodensity encoder is learned jointly through the auxiliary objective.

\subsubsection{\textbf{Overall objective.}}
\textsc{X-CGM-JEPA} combines the CGM predictive loss with the cross-view Glucodensity loss:
\[
\mathcal{L}_{\text{total}}=\mathcal{L}_{\text{CGM}}+\lambda\,\mathcal{L}_{\text{GD}},
\]
where $\lambda$ is a nonnegative coefficient that balances the cross-view Glucodensity objective against the CGM predictive objective. We fix $\lambda{=}1$ in the main experiments and analyze sensitivity to $\lambda$ in the ablation study.

\subsection{Downstream Evaluation Protocol}
After pretraining, we freeze the CGM encoder $f_\theta$ and extract embeddings for labeled subjects. We then train simple a linear classifier head (logistic regression) on top of embeddings for both insulin resistance $\beta$-cell dysfunction. We report AUROC, PRAUC, and F1 under multiple evaluation settings, including controlled in-clinic venous evaluation, venous-to-CGM cross-modality transfer, and in-domain home CGM evaluation.

\section{Dataset Overview}
\label{sec:app:data}

We use two complementary CGM corpora: an unlabeled corpus assembled for self-supervised pretraining and a labeled corpus, with paired venous and continuous-glucose measurements, for downstream metabolic-subphenotype evaluation.

\subsection{Pretraining Sensor Dataset}
\label{sec:app:data:pretrain}
The pretraining corpus pools two open CGM cohorts into a single unlabeled stream of 5-minute glucose readings:
\begin{Itemize}
    \item \textbf{Stanford initial cohort.} 22 subjects from the Stanford CGM study \citep{metwally2025prediction} who carry continuous CGM data in the released file (the remaining 5 subjects of the initial cohort have only OGTT measurements and are excluded).
    \vspace{3pt}
    \item \textbf{Colas CGM \citep{colas2019detrended}.} 206 subjects whose recordings are sliced into full-day windows, yielding 391 daily samples.
\end{Itemize}
After cohort merging, the pretraining table contains \textbf{413 subject-days} ($\approx$\,389{,}365 rows at 5-minute resolution). Streams are segmented into 24-hour windows of length 288, with optional sliding stride (288 = no overlap; 144 = 50\% overlap). All windows are tokenized into $P{=}24$ non-overlapping hourly patches of size 12 prior to encoding. Glucodensity views used by \textsc{X-CGM-JEPA} are pre-computed once per window via Gaussian KDE on a $32{\times}32$ grid and patchified spatially with patch size 8. No subject-level normalization is applied to the input glucose values in the default configuration.

\subsection{Downstream Datasets}
\label{sec:app:data:downstream}
Downstream evaluation uses the labeled subset of the Stanford CGM study \citep{metwally2025prediction} with metabolic phenotypes derived from gold-standard clinical assays:
\begin{Itemize}
    \item \textbf{Insulin Resistance (IR).} Binary label derived from \texttt{sspg\_2\_classes} (steady-state plasma glucose dichotomized into Insulin-Resistant vs.\ Insulin-Sensitive).
    \vspace{3pt}
    \item \textbf{$\beta$-cell dysfunction (Beta).} Binary label derived from \texttt{di\_2\_classes\_median} (median-split of the disposition index into Dysfunction vs.\ Normal).
\end{Itemize}
We adopt the Stanford-released cohort partitioning into two splits, distinguished by the presence of paired at-home CGM:
\begin{Itemize}
    \item \textbf{Initial cohort (train, $n{=}27$).} Subjects with \emph{venous OGTT only} (no matching CGM, no planned at-home CGM). Each entry contains a single \texttt{ctru\_venous} OGTT trace of 39 glucose values at 5-minute intervals spanning $t{=}{-}10$ to $t{=}180$ minutes, smoothed with a smoothing spline at $\lambda{=}0.35$ with internal $-1$ gaps interpolated.
    \vspace{3pt}
    \item \textbf{Validation cohort ($n{=}17$).} Subjects with \emph{paired venous OGTT and at-home CGM}, after removing dual-cohort overlaps with the initial cohort. Each entry exposes six aligned extraction methods: \texttt{ctru\_venous} (in-clinic venous), \texttt{ctru\_cgm} (in-clinic CGM during the OGTT), \texttt{home\_cgm\_1} and \texttt{home\_cgm\_2} (two at-home CGM sessions), \texttt{cgm\_home\_mean} (mean of the two home sessions), and \texttt{cgm\_all\_mean} (mean of \texttt{ctru\_cgm}, \texttt{home\_cgm\_1}, \texttt{home\_cgm\_2}). Smoothing uses $\lambda{=}0.4$.
\end{Itemize}
For reference, the exact \texttt{exp\_type} flag values used to filter the upstream Stanford release are:
\begin{center}
\small
\begin{tabular}{@{}lp{0.72\linewidth}@{}}
\toprule
\textbf{Cohort} & \textbf{\texttt{exp\_type} flag} \\
\midrule
Initial    & \texttt{venous\_without\_\allowbreak matching\_\allowbreak cgm\_\allowbreak and\_\allowbreak without\_\allowbreak planned\_\allowbreak athome\_\allowbreak cgm} \\
\addlinespace
Validation & \texttt{venous\_with\_\allowbreak matching\_\allowbreak cgm\_\allowbreak and\_\allowbreak with\_\allowbreak planned\_\allowbreak athome\_\allowbreak cgm} \\
\bottomrule
\end{tabular}
\end{center}
The combination of split type and extraction method defines the three evaluation pipelines reported in the main paper: in-clinic venous (\texttt{ctru\_venous}\,$\to$\,\texttt{ctru\_venous}), cross-modality transfer (\texttt{ctru\_venous}\,$\to$\, \texttt{home\_cgm\_mean}), and in-domain home CGM (\texttt{cgm\_home\_mean}\,$\to$\,\texttt{cgm\_home\_mean}).

\subsection{Sampling Rates and Preprocessing}
\label{sec:app:data:preproc}
The downstream cohort exposes two physically distinct acquisition modalities, \emph{venous OGTT} (clinical phlebotomy) and \emph{CGM} (subcutaneous interstitial sensor), which arrive at different cadences. We homogenize all streams onto a common 5-minute grid via a single shared alignment and smoothing pipeline, but the effect of that pipeline differs by modality, as we describe below.

\paragraph{Common alignment grid.} For every subject and every extraction method, we instantiate a 5-minute grid spanning $t = -10$ to $t = 180$ minutes (39 positions). For each grid position, we look up an exact-timepoint match in the source \texttt{(Timepoint, Glucose)} table; if no observation exists at that exact timepoint, the slot is filled with a sentinel value of \texttt{-1}. We deliberately do \emph{not} round neighboring timepoints onto the grid, so unaligned acquisitions remain marked as missing.

\paragraph{Venous OGTT: irregular and sparse.} The \texttt{ctru\_venous} stream consists of clinical phlebotomy draws at the Stanford Clinical Translational Research Unit during a standard oral glucose tolerance test (OGTT). Native acquisition times are the OGTT clinical sampling timepoints (typically $t \in \{-10, 0, 15, 30, 60, 90, 120, 150, 180\}$ minutes), so only $\sim$\,$9$ of the $39$ grid slots receive an observation; the remaining $\sim$\,$30$ slots are sentinel-filled. To recover a usable 5-minute trace, we fit a non-parametric smoothing spline via \texttt{scipy.interpolate.make\_smoothing\_spline} on the non-sentinel positions, then evaluate the fitted spline at \emph{all} 39 grid positions; sentinel positions therefore receive interpolated values on output. We use regularization $\lambda = 0.35$ for the initial cohort and $\lambda = 0.4$ for the validation cohort. Because the venous trace is sparse to begin with, the spline here is doing \emph{interpolation under denoising}, and the fitted curve is the only object the model ever sees from the venous channel.

\paragraph{CGM: natively 5-minute aligned.} CGM streams (\texttt{ctru\_cgm}, \texttt{home\_cgm\_1}, \texttt{home\_cgm\_2}) are acquired at the sensor's native 5-minute cadence and therefore land directly on the alignment grid: nearly all 39 positions receive a real observation, and very few sentinels are produced. We pass these streams through the same alignment and smoothing pipeline as the venous channel for a uniform interface; with the CGM streams the spline is effectively denoising only, since there are no large gaps to interpolate. Two derived ``mean'' streams used in our cross-modality experiments are pointwise averages over only the components that have a value at each grid position:
\begin{align*}
\texttt{cgm\_home\_mean}_t &= \mathrm{mean}\{\texttt{home\_cgm\_1}_t,\, \texttt{home\_cgm\_2}_t\}, \\
\texttt{cgm\_all\_mean}_t  &= \mathrm{mean}\{\texttt{ctru\_cgm}_t,\, \texttt{home\_cgm\_1}_t,\, \texttt{home\_cgm\_2}_t\}.
\end{align*}
A small number of subjects (e.g., \texttt{S80}) lack one of the at-home sessions; for those subjects the missing component is simply omitted from the mean rather than imputed. \texttt{cgm\_all\_mean} provides a single, denoised CGM trajectory per subject and is the canonical \emph{val\_extract\_method} for the in-domain home CGM evaluation in the main paper.

\subsection{Data Acquisition and Approval}
All data used in this work are derived from publicly released CGM corpora. The Stanford CGM study and its labeled metabolic subphenotypes are distributed via the \href{https://github.com/aametwally/Metabolic_Subphenotype_Predictor}{\texttt{Metabolic\_Subphenotype\_Predictor}} repository, and the Colas et al.\ CGM dataset is released by the original authors. Both sources retain their original IRB approvals and de-identification protocols; no additional human-subject data were collected for this work, and our preprocessing only operates on already de-identified glucose traces. We release the merged, preprocessed dataset (\texttt{Dataset\_Open/}) together with the train/validation split files used in our experiments to support reproducibility.

\section{Implementation Details}
\label{sec:app:implement}

\subsection{Model Architecture}
\label{sec:app:implement:arch}
The CGM context encoder $f_\theta$ is a lightweight Transformer over patch tokens. Each hourly patch of size 12 is mapped to an embedding via a 1D convolutional patch embedder with kernel size 3 and bias enabled. Embedded patches pass through a stack of standard pre-norm Transformer blocks, after which optional sinusoidal time features can be appended (disabled by default to avoid embedding mismatch with downstream sets that lack timestamps). The target encoder $f_{\bar{\theta}}$ shares the architecture of $f_\theta$ and is updated by an exponential moving average (EMA) of $f_\theta$. The masked-patch predictor $p_\phi$ is a smaller Transformer operating on positional-conditioned context tokens.

\begin{table}[h]
\centering
\small
\begin{tabular}{lcc}
\toprule
\textbf{Component} & \textbf{Hyperparameter} & \textbf{Value} \\
\midrule
Patch embedder      & patch size / conv kernel        & 12 / 3 \\
Context encoder     & embed dim / heads / layers      & 96 / 6 / 3 \\
Context encoder     & embed bias / dropout            & True / 0.0 \\
Target encoder      & EMA momentum                    & 0.997 \\
Predictor           & embed dim / heads / layers      & 48 / 2 / 1 \\
Glucodensity        & KDE grid / spatial patch size   & 32$\times$32 / 8 \\
Window              & length / patches per window     & 288 / 24 \\
\bottomrule
\end{tabular}
\caption{\textsc{CGM-JEPA} / \textsc{X-CGM-JEPA} architecture defaults.}
\label{tab:app:arch}
\end{table}

\subsection{Pretraining Details}
We pretrain on day-windowed CGM with patch-level masking. The default mask ratio is $0.25$ (varied in the ablation study), and the cross-view weight $\lambda$ is fixed to $1.0$ in the main results. Optimization uses Adam with learning rate $10^{-4}$, an exponential learning-rate schedule (\texttt{step\_size}\,$=$\,100, \texttt{gamma}\,$=$\,0.99), a warmup ratio of $0.15$, gradient norm clipping at $1.0$, and an iteration-per-epoch scaling factor $\texttt{ipe\_scale}{=}1.25$. We train for 100 epochs with batch size 128 and seed 43. Mixed precision is not used; all runs fit on a single GPU.

\paragraph{Glucodensity view computation.}
\label{sec:app:implement:glucodensity}
The Glucodensity view used by \textsc{X-CGM-JEPA} compresses one daily CGM window into an image-like distributional summary that captures \emph{joint} structure between glucose level, glucose velocity, and glucose acceleration, and is therefore complementary to the time-domain signal seen by $f_\theta$. Given a daily CGM window $g_{1:288}$ we map indices to a continuous time axis $t = 5i / 60$ in hours and fit a non-parametric smoothing spline $\tilde{g}(t)$ via \texttt{scipy.interpolate.UnivariateSpline} with smoothing factor $1.0$. From this smoothed curve we obtain three pointwise channels:
\[
G(t)=\tilde{g}(t), \qquad \dot{G}(t)=\frac{d\tilde{g}}{dt}, \qquad \ddot{G}(t)=\frac{d^2\tilde{g}}{dt^2},
\]
i.e., level, speed, and acceleration. We then compute three two-dimensional Gaussian kernel-density estimates for the channel \emph{pairs} $(G,\dot{G})$, $(G,\ddot{G})$, and $(\dot{G},\ddot{G})$. Each KDE is evaluated on a $32 \times 32$ regular grid spanning the empirical 1\textsuperscript{st}--99\textsuperscript{th} percentile range of the corresponding channel pair (this trimming makes the grid robust to occasional spline-derivative spikes), and is normalized by its maximum density value to lie in $[0, 1]$. Stacking the three normalized KDE images along the channel axis gives a single 3-channel ``Glucodensity image'' $\mathbf{D} \in \mathbb{R}^{32 \times 32 \times 3}$ per window. Finally, we tile $\mathbf{D}$ into non-overlapping spatial patches of size $8\times 8$ in a Vision-Transformer-style layout, producing $4 \times 4 = 16$ patches that each carry all three channels jointly (i.e., each patch has shape $8 \times 8 \times 3$). Each patch is flattened into a $192$-dimensional token, yielding a sequence of $16$ tokens that is fed to the Glucodensity encoder $g_\psi$.

\paragraph{Pre-computation and caching.}
The KDE step is by far the most expensive part of the data pipeline: a fresh KDE evaluation per (window, pair, grid) tuple recomputes the same density tensor at every gradient step, dominating wall time even when the rest of the model is small. Because the Glucodensity view is a deterministic function of the daily CGM window (no augmentation, no stochasticity, no learnable parameters upstream), we instead precompute it once over the full pretraining corpus before training begins (see \texttt{utils/precompute\_glucodensity.py}). The precompute job iterates over the same windowed dataset that pretraining uses, calls the procedure described above for every $(subject, split\_idx)$ pair, and stores the resulting patches in a single keyed pickle cache together with the configuration ($\texttt{gridsize}=32$, $\texttt{spatial\_patch\_size}=8$, $\texttt{patch\_size}=12$, $\texttt{series\_split\_size}=288$). At training time, the data loader looks up the cached Glucodensity patches by sample key, eliminating KDE evaluation from the training loop entirely; eight CPU workers ($\texttt{gluco\_kde\_workers}=8$) are used only during precomputation. Precomputation runs once per dataset version on commodity CPU hardware and the resulting cache is checkpointed to disk and reused across all subsequent runs.

\subsection{End-to-End Pipelines}
\label{sec:app:implement:pipelines}
For clarity, we summarize the full data flow at training time and at evaluation time. Both pipelines share the preprocessing stack of Section~\ref{sec:app:data:preproc}.

\paragraph{Pretraining pipeline.}
\begin{enumerate}
    \item \textbf{Load} the pooled pretraining CSV (\texttt{cgm\_initial\_cohort.csv}, 22 Stanford + 206 Colas subjects, $\approx$\,389k rows at 5-minute cadence).
    \item \textbf{Slice} each subject's stream into 24-hour windows of length 288.
    \item \textbf{Tokenize} each window into $P{=}24$ non-overlapping hourly patches of size 12.
    \item \textbf{(\textsc{X-CGM-JEPA} only)} Look up the precomputed Glucodensity tensor for the same window from the pickle cache (Section~\ref{sec:app:implement:glucodensity}); no live KDE.
    \item \textbf{Mask} a random subset of patches at the configured mask ratio (default 0.25, varied in the ablation), splitting the window into context (visible) and target (masked) sets.
    \item \textbf{Forward.} Context encoder $f_\theta$ encodes visible patches; EMA target encoder $f_{\bar\theta}$ encodes the full window and provides the target latents at masked positions; predictor $p_\phi$ predicts the target latents from the context. For \textsc{X-CGM-JEPA}, an auxiliary cross-view predictor $q_\omega$ predicts the masked Glucodensity embedding $\mathbf{u}=g_\psi(\mathbf{D}^{(t)})$ from the same context.
    \item \textbf{Backward.} Total loss $\mathcal{L}_{\text{CGM}} + \lambda\,\mathcal{L}_{\text{GD}}$ (with $\mathcal{L}_{\text{GD}}{=}0$ for vanilla \textsc{CGM-JEPA}); Adam step on $\theta$, $\phi$, $\omega$, $\psi$; $\bar\theta$ is updated by EMA of $\theta$ at momentum 0.997. Gradients are stop-gradiented on the CGM target branch and \emph{flow} into $g_\psi$.
    \item \textbf{Checkpoint.} After 100 epochs, the encoder $f_\theta$ is logged as a versioned wandb artifact together with all metadata (mask ratio, $\lambda$, seed) so it is fully reproducible at downstream time.
\end{enumerate}

\paragraph{Downstream pipeline (linear probe).}
\begin{enumerate}
    \item \textbf{Resolve} a pretrained encoder by its wandb artifact version. Architecture and tokenizer dimensions are read from the artifact metadata, so the loader can reconstruct the model without recourse to a separate checkpoint config file.
    \item \textbf{Load} the labeled subject set from \texttt{train\_split.json} (initial cohort, $n{=}27$) and \texttt{validation\allowbreak\_split.json} (validation cohort, $n{=}17$); each subject's daily window is already on the 5-minute grid via Section~\ref{sec:app:data:preproc}.
    \item \textbf{Freeze} $f_\theta$; route each labeled window through the same patchify $\to$ encode pipeline used at pretraining time, then mean-pool the patch tokens to obtain a fixed-dimensional subject embedding $\mathbf{z}\in\mathbb{R}^{d}$ ($d{=}96$ for the encoders in this paper).
    \item \textbf{Probe.} Fit an $\ell_2$-regularized logistic-regression probe (\texttt{LogisticRegressionCV}) with class-balanced weighting and inner 2-fold CV over the regularization grid $C \in \{10^{-3}, 10^{-2}, 10^{-1}, 1, 10, 100\}$, scored by AUROC. Outer 2-fold cross-validation is repeated for 20 random seeds, giving $20{\times}2{=}40$ paired test folds per \emph{(encoder, task)} cell.
    \item \textbf{Evaluate.} Report mean AUROC, PR-AUC, and F1 across folds and use a paired bootstrap test for headline comparisons. Optional probes (Linear SVC, Ridge, Random Forest, kNN) follow the same outer protocol.
\end{enumerate}
The \emph{evaluation pipeline} is parameterized by an \texttt{(extract\_method, val\_extract\_method)} pair: \emph{in-clinic venous} (\texttt{ctru\_venous}\,$\to$\,\texttt{ctru\_venous}) trains and tests on the OGTT venous trace; \emph{cross-modality transfer} (\texttt{ctru\_venous}\,$\to$\, \texttt{cgm\_home\_mean}) trains on venous and tests on CGM, evaluating modality robustness; \emph{in-domain home CGM} (\texttt{cgm\_home\_mean}\,$\to$\,\texttt{cgm\_home\_mean}) restricts both training and testing to the at-home CGM mean.

\subsection{Self-Supervised Learning Baselines}
\label{sec:app:implement:ssl_baselines}
We compare \textsc{CGM-JEPA} and \textsc{X-CGM-JEPA} against two SSL baselines, each pretrained from scratch on the same pooled CGM corpus and evaluated under the identical linear-probe pipeline of the previous subsection. To control for capacity differences, we deliberately match the baseline encoder sizes to the size of our own context encoder; see the per-baseline notes below.

\paragraph{TS2Vec \citep{yue2022ts2vecuniversalrepresentationtime}.}
\textbf{Pretraining objective.} Hierarchical contrastive representation learning for time series, where positive pairs are obtained by random cropping of the same series and the contrastive loss is aggregated across multiple temporal scales via dilated convolutions.

\textbf{Architecture.} A stack of $10$ dilated 1D-convolutional residual blocks (\texttt{depth}\,$=$\,10) with hidden width 64 producing a representation of dimension 96 (matched to ours), no Transformer attention. Input is the raw univariate glucose stream (no patchification, no tokenization).

\textbf{Pretraining hyperparameters.} Learning rate $10^{-3}$ (Adam, default), batch size 128, maximum input length 3000 timesteps, $\texttt{temporal\_unit}{=}0$ (contrast at the finest temporal scale), 100 epochs.

\textbf{Pipeline notes.} TS2Vec's reference implementation expects \texttt{NaN} to mark missing values, so we convert our \texttt{-1} sentinels (Section~\ref{sec:app:data:preproc}) to \texttt{NaN} on the fly. We also enable z-score normalization at the dataset-statistics step ($\texttt{normalize\_x}{=}\textbf{True}$ when computing the loader's mean/std), following the original TS2Vec recipe; this is the only baseline where we deviate from the otherwise-default \texttt{normalize\_x}\,$=$\,False, because the contrastive loss is sensitive to absolute-scale drift across subjects.

\paragraph{GluFormer \citep{lutsker2025gluformer}.}
\textbf{Pretraining objective.} Masked-token prediction over a discrete glucose vocabulary, analogous to BERT-style masked language modeling but on glucose tokens.

\textbf{Tokenization.} Continuous glucose values are quantized into $K{=}280$ uniformly spaced bins between the empirical min and max of the corpus, plus one extra index reserved for padding (vocabulary size $K{+}1{=}281$). The bin width is therefore $(g_{\max}-g_{\min})/280$\,mg/dL.

\textbf{Architecture (size-matched).} \emph{We deliberately reduce GluFormer below the configuration in the original paper}: token embedding dim $96$, $6$ attention heads, $3$ encoder layers, feed-forward width $192$ ($=2{\times}\text{embed dim}$), maximum sequence length 25{,}000 tokens (sufficient to cover the longest sliding windows produced under stride 144), dropout $0.0$. The original GluFormer architecture is much wider/deeper and was developed for substantially larger CGM corpora; on our pooled corpus of $\approx$\,400 subject-days a full-size GluFormer would dominate the dataset and overfit, so we shrink it to match our own context encoder so that any SSL-vs-SSL difference reflects the \emph{objective} rather than parameter-count.

\textbf{Pretraining hyperparameters.} Adam with learning rate $10^{-4}$, exponential LR schedule (\texttt{step\allowbreak\_size}\,$=$\,100, \texttt{gamma}\,$=$\,0.99) matching our \textsc{CGM-JEPA} run, gradient clipping at norm 1.0, batch size 128, 101 epochs, seed 43. The masking rate matches the value reported in the original paper (default: 25\%); padded positions are excluded from the cross-entropy loss.

\subsection{TSFM Baselines}
\label{sec:app:implement:gemini_baselines}
For zero-shot probing of generic time-series foundation models (TSFMs), we compare against two off-the-shelf TSFMs without any glucose-specific pretraining or finetuning. In both cases we follow the standard inference recipes published by the original authors for downstream classification, restricted to the inputs we use elsewhere in this paper.

\paragraph{Mantis \citep{feofanov2025mantislightweightcalibratedfoundation}.}
\textbf{Checkpoint.} \texttt{paris-noah/Mantis-8M}, the publicly released $8$M-parameter \texttt{Mantis8M} model loaded via the official \texttt{mantis} Python package (\texttt{Mantis8M.from\_pretrained(\dots)} wrapped in a \texttt{MantisTrainer}, exactly as in the project's reference scripts).

\textbf{Inference protocol.} Mantis was trained on inputs of length $512$, and the official \texttt{MantisTrainer.\allowbreak transform} API expects this length verbatim. Since our daily CGM window is 288 steps, we follow the standard practice from the released code base and \emph{linearly interpolate} the window up to length 512 (\texttt{F.interpolate(\dots, mode="linear")}) before passing it to the encoder; the channel axis is set to $1$ for univariate input, giving a final shape $(B, 1, 512)$. The model is run under \texttt{torch.no\_grad()} and returns a single fixed-dimensional embedding per window, which we use directly as the downstream feature without any additional pooling.

\paragraph{MOMENT \citep{goswami2024moment}.}
\textbf{Checkpoints.} We evaluate both released sizes, \texttt{AutonLab/MOMENT-1-small} and \texttt{AutonLab/MOMENT-1-large}, loaded through the official \texttt{momentfm} package as \texttt{MOMENTPipeline.\allowbreak from\_pretrained}.

\textbf{Inference protocol.} We initialize each pipeline with the kwargs the MOMENT authors prescribe for \emph{embedding}-mode usage in their classification tutorial.\footnote{\href{https://github.com/moment-timeseries-foundation-model/moment/blob/main/tutorials/classification.ipynb}{\texttt{moment-timeseries-foundation-model/moment} -- \texttt{tutorials/classification.ipynb}}} concretely \texttt{model\_kwargs=\{"task\_name":\allowbreak "embedding", "n\_channels":1\}}, followed by \texttt{model.init()} and \texttt{model.eval()}. MOMENT also expects an input length of $512$; rather than interpolating, we follow MOMENT's recommended approach for short series and \emph{left-pad} our 288-step daily window with zeros to length 512 and pass an explicit \texttt{input\_mask} (1 at real positions, 0 at padded positions) so the encoder ignores the synthetic prefix. The pipeline is called as \texttt{output = pipeline(x\_enc=x, input\_mask=input\_mask)} under \texttt{torch.no\_grad()} and we use \texttt{output.embeddings} directly as the downstream feature; no additional pooling is applied.

\paragraph{Common to both TSFMs.} Both encoders are frozen at their published weights; no fine-tuning, prompt-tuning, or domain adaptation is performed. The TSFM-derived embeddings are fed to the same linear probe described in Section~\ref{sec:app:implement:pipelines}, so the only difference between rows in the TSFM comparison is the upstream encoder. Any gap between TSFM rows and our \textsc{CGM-JEPA} / \textsc{X-CGM-JEPA} rows therefore isolates the value of CGM-specific pretraining rather than reflecting evaluation-pipeline asymmetries.

\section{Societal Impact}
\label{sec:app:impact}

\paragraph{\textbf{Broader impacts.}}
Type 2 Diabetes (T2D) affects over $537$ million adults globally, making early detection a public-health priority. The two principal pathways toward T2D, insulin resistance and $\beta$-cell dysfunction, implicate different lifestyle and therapeutic responses, so distinguishing them, not merely flagging overall risk, is what makes early intervention actionable. Yet the gold-standard test for these subphenotypes, the venous oral glucose tolerance test, is invasive and resource-intensive, ruling it out for population-scale screening. By showing that self-supervised representations from unlabeled CGM data support clinically meaningful subphenotype discrimination under realistic deployment conditions, our work contributes a foundation for non-invasive, scalable metabolic risk stratification using consumer-grade wearables, lowering the barrier to targeted early prevention.

\paragraph{\textbf{Limitations and Ethical Considerations.}}
While population-scale CGM-based screening holds great promise for early metabolic risk stratification, it must be guided by careful attention to safety, fairness, and privacy. The possibility of misuse by malicious actors underscores the importance of responsible development. Although we advocate for open science and release code, training configurations, and pretrained weights, health data requires a delicate balance between reproducibility and participant confidentiality: we release de-identified CGM only from participants who consented to data sharing, while the full clinical cohorts remain access-controlled.

Importantly, \textsc{CGM-JEPA}'s family is a research prototype and is not intended for clinical use. It has not been validated as a diagnostic tool and should not be used for medical decision-making without formal regulatory approval. While our subgroup analysis (Section~\ref{subsec:subgroup-redistribution}) shows the cross-view objective compresses worst-to-best AUROC gaps across demographic strata, per-subgroup sample sizes are small ($n = 5$--$12$) and these patterns are not yet validated beyond our cohorts; deployment without re-evaluation could produce inequitable outcomes our analysis does not predict. Clinical deployment would require prospective multi-site validation and compliance with the relevant medical-device regulatory framework.

Finally, while \textsc{CGM-JEPA}'s family methodology is designed to be generalizable, our evaluation is limited to two cohorts ($N{=}27$ and $N{=}17$ in the public-release subset), one CGM device, and OGTT-derived labels for two subphenotypes. Further research is needed across broader device ecosystems, longer temporal windows, additional metabolic outcomes, and more diverse populations.

\end{document}